\documentclass{article}

    \PassOptionsToPackage{numbers, compress}{natbib}

     \usepackage[preprint]{neurips_2020}

\usepackage[utf8]{inputenc} 
\usepackage[T1]{fontenc}    
\usepackage{hyperref}       
\usepackage{url}            
\usepackage{booktabs}       
\usepackage{amsfonts}       
\usepackage{nicefrac}       
\usepackage{microtype}      
\usepackage{caption}
\usepackage{subcaption}
\usepackage{graphicx}
\usepackage{mathtools,amssymb}
\usepackage{amsmath}
\newcommand{\norm}[1]{\left\lVert#1\right\rVert}
\usepackage[linesnumbered,ruled,vlined]{algorithm2e}
\usepackage{algpseudocode}
\usepackage{comment}
\usepackage{soul}
\usepackage{color}
\usepackage{multirow}
\usepackage{colortbl}
\usepackage[table]{xcolor}

\usepackage{color,array}

\usepackage[locale=FR]{siunitx}
\usepackage{booktabs}
\usepackage[table]{xcolor}
\usepackage{caption}
\usepackage{subcaption}
\usepackage{mathtools,amssymb}
\usepackage{amsmath}
\usepackage[linesnumbered,ruled,vlined]{algorithm2e}
\usepackage{algpseudocode}
\usepackage{comment}
\usepackage{enumitem}
\usepackage{soul}
\usepackage{color}
\usepackage{pgfplots}
\pgfplotsset{compat=1.15}
\usepackage{tikz}
\usepackage{tikz-qtree}
\usepackage{pgfplots}
\usetikzlibrary{patterns}
\DeclareMathOperator*{\argmax}{arg\,max}

\usepackage{amsthm}

\newtheorem{theorem}{Theorem}
\usepackage{multirow}
\usepackage{booktabs}
\usepackage{graphicx}
\usepackage{caption}
\usepackage{tabularx}

\newtheorem{propositions}{Proposition}

\usepackage{makecell}
\usepackage{wrapfig}

\title{Adversary Agnostic \\Robust Deep Reinforcement Learning}

\author{%
  Xinghua Qu \\
  School of Computer Science \& Engineering\\
  Nanyang Technological University\\
  \texttt{xinghua001@e.ntu.edu.sg} \\
   \And
  Yew-Soon Ong \\
  School of Computer Science \& Engineering\\
  Nanyang Technological University \\
  \texttt{asysong@ntu.edu.sg} \\
  \And
  Abhishek Gupta \\
  Singapore Institute of Manufacturing Technology\\
Agency for Science, Technology and Research\\
  \texttt{abhishek\_gupta@simtech.a-star.edu.sg} \\
  \And
  Zhu Sun \\
  Department of Computing \\
   Macquarie University \\
   \texttt{sunzhuntu@gmail.com} \\
}
\pgfplotsset{compat=1.14}
\begin{document}
\maketitle
\begin{abstract}
Deep reinforcement learning (DRL) policies have been shown to be deceived by perturbations (e.g., random noise or intensional adversarial attacks) on state observations that appear at test time but are unknown during training. 
To increase the robustness of DRL policies, previous approaches assume that the knowledge of adversaries can be added into the training process to achieve the corresponding generalization ability on these perturbed observations. 
However, such an assumption not only makes the robustness improvement more expensive, but may also leave a model less effective to other kinds of attacks in the wild.
In contrast, we propose an \textit{adversary agnostic} robust DRL paradigm that does not require learning from adversaries.
To this end, we first theoretically derive that robustness could indeed be achieved independently of the adversaries based on a policy distillation setting. Motivated by this finding, we propose a new policy distillation loss with two terms:
1) a prescription gap maximization loss aiming at simultaneously maximizing the likelihood of the action selected by the teacher policy and the entropy over the remaining actions; 2) a corresponding Jacobian regularization loss that minimizes the magnitude of gradient with respect to the input state. The theoretical analysis shows that our distillation loss guarantees to increase the prescription gap and the adversarial robustness. Furthermore, experiments on five Atari games firmly verify the superiority of our approach in terms of boosting adversarial robustness compared to other state-of-the-art methods. 
\end{abstract}

\section{Introduction}
The advancements in deep reinforcement learning (DRL) have demonstrated that deep neural networks (DNNs) as powerful function approximators can be trained to prescribe near-optimal actions on many complex tasks (e.g., Atari games \cite{mnih2015human}, robotics control \cite{kormushev2013reinforcement} and motor control \cite{traue2020toward}). Although remarkable achievements have been documented, many studies on adversarial attacks \cite{huang2017adversarial, lin2017tactics, qu2020minimalistic, xiao2019characterizing} have shown that DRL policies can be easily deceived. This inspires the studies on improving the adversarial robustness of DRL policies, so as to defend against the adversarial attacks.

To enhance the adversarial robustness of DRL policies, many studies have investigated adding the adversaries into the training process to achieve the corresponding generalization ability on these perturbed observations. For instance, Mirman et al.~\cite{mirman2018distilled} and Fischer et al.~\cite{fischer2019online} utilized adversarial training \cite{pinto2017robust} based policy distillation \cite{rusu2015policy} to obtain an accurate and robust student policy, where the robustness is learned from the added adversarial attacks that are generated by fast gradient sign method and projected gradient descent, respectively. 
Recently, based on a derivation of state-adversarial Markov decision process (SA-MDP), Zhang et al. \cite{zhang2020robust} proposed a robustness regularization to align the policy's prescriptions \textit{with} and without adversarial perturbation; thus still being adversary dependent.
In a nutshell, these approaches require additional procedures of incorporating adversaries in the training, {which, however, makes improvements on robustness less flexible}
({e.g., robust policy depending on one particular attack may fail when another attack happens}), and more expensive from both computation (in calculating the adversaries) and implementation (in deploying adversaries for training in the wild) perspectives. 

Instead of learning robust policy from adversaries, under the policy distillation (PD) paradigm, we prove in Section 3.2 that such robustness can also be achieved without relying on adversaries. Specifically, a robust student policy can be learned through maximizing the student policy's prescription gap between the teacher selected action and the remaining actions under attack. Most importantly, we further derive that maximizing the prescription gap under attack can be transformed to simultaneously maximizing the prescription \textbf{without attack} and minimizing the Jacobian with respect to input states; this provides us the possibility of achieving adversarial robustness without learning from adversaries. 

Inspired by this derivation, we propose an \textit{adversary agnostic PD (A2PD)} loss that includes two parts: 1) prescription gap maximization (PGM) loss, and 2) Jacobian regularization (JR) loss. The PGM loss is different from most previous distillation loss functions (e.g., cross-entropy) that merely maximize the probability of the action selected by teacher policy. In contrast, we also maximize the entropy of those actions not selected by the teacher policy, which enforces the student policy to have a larger prescription gap in order to resist perturbations in state observations. More importantly, the entropy term is weighted by the probability of the selected action; this allows the training to focus on PD accuracy at the beginning and pursue entropy maximization in the end. Last but not the least, in order to further improve the adversarial robustness, we also minimize the magnitude of Jacobian with respect to the input state, which is calculated based on PGM loss.

\textbf{Our main contributions can be summarized as: }
\begin{itemize}[leftmargin=*]
    \item For the first time in DRL, we propose the adversary agnostic robust DRL that achieves robustness without relying on learning from adversaries. To this end, based on the PD paradigm, we theoretically derive that the robustness of student policy can be indeed learned independently of the adversaries. 
    \item To this end, we design the adversary agnostic PD (A2PD) loss function that contains: 1) a PGM loss for simultaneously maximizing the probability of the action prescribed by teacher policy as well as the entropy of unwanted actions; 2) a JR loss that minimizes the norm of Jacobian with respect to the input state.
    \vspace{-0.05in}
    \item The theoretical analysis proves that our proposed PD loss guarantees to increase the prescription gap and the adversarial robustness. Meanwhile, experiments on five Atari games show that the robustness of the student policies trained by A2PD loss is significantly improved.
\end{itemize}


\section{Related Work}
In the context of DRL, Huang et al.~\cite{huang2017adversarial} were among the first to analyze the vulnerability of DNN policies, where they utilized the fast gradient sign method (FGSM)~\cite{goodfellow2015explaining} to generate adversarial perturbations. Lin et al.~\cite{lin2017tactics} explored a more complicated scenario by partially perturbing only selected frames, and they also investigated a designated targeted attack using a generative model. Qu et al.~\cite{qu2020minimalistic} studied a minimalistic attack to showcase that merely perturbing a single pixel in a few selected frames can significantly degrade the reward of state-of-the-art policies. Besides, Xiao et al.~\cite{xiao2019characterizing} provided a survey that refers many other attacks on RL with different settings.

To resist against adversarial attacks in DRL, there have been several works that study the adversarial robustness improvement by using adversarial training \cite{yuan2019adversarial}. Mandlekar et al.~\cite{mandlekar2017adversarially} applied adversarial training on policy gradient algorithm by leveraging a simple FGSM to generate adversarial examples, but they just tested on some simple RL tasks (i.e, Mujoco locomotion with relatively low dimensional input state). Pattanaik et al.~\cite{pattanaik2018robust} introduced much stronger attacks that are achieved by projected gradient descent (PGD) in adversarial training on Atari games. However, the results showcase that the robustness increase causes significant performance drop. To obtain better robustness, Mirman et al.~\cite{mirman2018distilled} and Fischer et al.~\cite{fischer2019online} proposed adversarial training based policy distillation
to build a more robust student policy, where FGSM and PGD are utilized respectively to generate adversarial attacks during training.
Recently, Zhang et al. \cite{zhang2020robust} proposed a robustness regularization to align the policy's prescriptions {with} and without adversarial perturbation; the robustness improvement is therefore still dependent on adversaries.

In sum, those approaches require additional procedures to operate adversaries during training; this makes improvements on adversarial robustness less flexible and more expensive in computation and real-world implementation. In contrast, our policy distillation approach is able to learn a robust student policy that does not rely on any information of the adversaries. The \textit{\textbf{broader impact}} of our adversary agnostic approach for improving adversarial robustness would be highlighted in those safety critical applications, since our robustness can be achieved without requiring the expensive and dangerous adversarial examples to be involved in the training process. For instance, in autonomous driving, previous adversary dependent approaches may needs to witness traffic accidents (as the results of perturbed action selection) with the goal of improving the robustness against input perturbations. \textit{In contrast}, our A2PD totally gets rid of those dangerous and un-affordable data. 
Therefore, this paper provides a more realistic solution for improving the adversarial robustness of DRL in the wild.

\section{Methodology}
In this section, we start with providing preliminaries on deep reinforcement learning and policy distillation (PD). Based on PD paradigm, we theoretically derive how to achieve a robust student policy in resisting against adversaries but without relying on adversaries. Inspired by this derivation, we thereby propose a novel policy distillation loss $\mathcal{L}_{A2PD}$, consisting of a prescription gap maximization loss and a Jacobian regularization loss. 
Finally, we theoretically prove that our distillation loss can increase the prescription gap and the adversarial robustness. 

\subsection{Preliminary}
\textbf{Deep Reinforcement Learning (DRL).} In this paper, we consider a finite-horizon Markov decision process (MDP) that consists of a 4-tuple $(\mathcal{S},\mathcal{A}, r,p)$, where $\mathcal{S}$ denotes the state space; $\mathcal{A}$ means the action space with size $|\mathcal{A}|$; $r(s_t)$ is the reward function when state $s_t$ transits to $s_{t+1}$ given action $a_t$; and $p$ represents the state transition function, e.g., $p(s_{t+1}|s_t,a_t)$, that is controlled by the environment. The aim of RL\footnote{In this paper, DRL and RL are interchangeably used.} algorithm (e.g., DQN~\cite{mnih2015human}) is to maximize the expected accumulative reward $\mathcal{R}(\pi_{\theta})=\mathbb{E}[\sum\nolimits_{t=0}^{T}\gamma^t r(s_t)|\pi_{\theta}]$ following a policy $\pi_{\theta}$, where $\pi$ is parameterized by $\theta$; $\gamma$ is the discount factor. However, $\mathcal{R}(\pi_{\theta})$ can be significantly degraded when an adversarial example $\delta_t: \mathcal{S}\rightarrow \mathcal{S}$ exists in state $s_t$. Note that in this paper $\pi_{\theta}(s_t)$ represents a prescribed distribution in action space of the policy $\pi_{\theta}$ on state $s_t$; $\pi_{\theta}(s_t, a)$ is the prescription on action $a$ given policy $\pi_{\theta}$ and state $s_t.$ In the adversarial attack setting, $\delta_t$ is added on the original state $s_t$ in order to perturb the prescribed action distribution $\pi_\theta(s_t+\delta_t)$. Therefore, the perturbed action $a_t=\argmax_a\pi_\theta(s_t+\delta_t, a)$ may be sub-optimal, thus reducing the reward of $\pi_\theta$. The expected accumulative reward with perturbation $\delta_t$ is denoted as $\mathcal{R}(\pi_{\theta})=\mathbb{E}[\sum\nolimits_{t=0}^{T}\gamma^t r(s_t,\delta_t)|\pi_{\theta}]$. To improving the robustness in resisting against $\delta_t$, adversarial training based policy distillations~\cite{fischer2019online} have been used.

\textbf{Policy Distillation (PD).} We follow the problem setting of PD~\cite{rusu2015policy}, where a teacher policy $\pi_{\theta^T}$ (e.g., $Q$ value approximator) is first learned by RL algorithms. The aim of PD is to learn a student policy $\pi_{\theta^S}$ that can mimic the behavior of its teacher policy $\pi_{\theta^T}$. 
Therefore, PD is formulated to minimise the loss function $\mathcal{L}(\theta^S)$ that measures the difference between the prescription 
from student policy $\pi_{\theta^S}(s_t)$ and that from the pre-trained teacher policy $\pi_{\theta^T}(s_t)$, which is shown as 


\begin{equation}
     \mathcal{L}(\theta^S) = \mathbb{E}_{{s_t} \sim \mathcal{S}}\left[\mathcal{D}(\pi_{\theta^S}(s_t), \pi_{\theta^T}(s_t))\right],
\end{equation}
where $\mathcal{D}$ is a distance measurement and it usually adopts the Kullback–Leibler (KL) divergence~\cite{czarnecki2019distilling} or cross entropy~\cite{fischer2019online}.

Although PD has documented many success stories on reward improvement~\cite{rusu2015policy} and policy network compression~\cite{czarnecki2019distilling}, the adversarial robustness of the student policy $\pi_{\theta^S}$ has been less investigated so far. Recent advancements (e.g., Fischer et al.~\cite{fischer2019online}) have studied the adversarial defense by involving adversarial training in PD. In doing so, the loss function for a robust student policy $\mathcal{L}_R(\theta^S)$ is reformulated as,
\begin{equation}
\small
    \mathcal{L}_R(\theta^S) = \mathbb{E}_{{s_t} \sim \mathcal{S}}\left[\max_{\delta_t}\mathcal{D}\left(\pi_{\theta^S}(s_t+\delta_t), \pi_{\theta^T}(s_t)\right)\right], \text{  }\norm{\delta_t}\leqslant\epsilon ,
\end{equation}
where the norm value of $\delta_t$ is bounded by $\epsilon$. In order to generate adversarial examples $\delta_t$, many attack models (e.g., FGSM \cite{mirman2018distilled} and PGD \cite{fischer2019online}) have been applied. In particular, Zhang et al.~\cite{zhang2020robust} proposed the SA-MDP with a  robustness regularization to align the  policy’s prescriptions with and without adversarial perturbation; thus the regularization is still dependent on the adversaries $\delta_t$. In summary, these approaches for robustness require accessing and operating on the adversaries; this makes adversarial robustness improvement less flexible. That is to say, the robustness obtained depending on one particular attack may fail when the agent faces another attack. Furthermore, the generation of adversarial examples may also lead to higher computational cost. Hence, a natural question to ask is: can we build a distillation paradigm that is capable of improving robustness without learning from adversaries in the training of student policy?

\subsection{Adversarial Robustness without Adversaries in DRL}
The aim of our policy distillation is to find a student policy $\pi_{\theta^S}$ that can maximize the accumulated reward $\mathcal{R}(\pi_{\theta^S})$ even with adversarial perturbation $\delta_t$ on state $s_t$, while the distillation training is independent on adversaries. With this in mind,  we note that the expected reward starting from $s_t$ is denoted by state value $V(s_t)$:
\begin{equation}
\small
    V(s_t) = \mathbb{E}\left[\sum\nolimits_{k=1}^{T}\gamma^k r(s_{t+k})|\pi_{\theta^S}\right],
\end{equation}
where $\gamma$ is the discount factor. Thus, given adversarial perturbation $\delta_t$ on state $s_t$, the adversarial state value $ V(s_t+\delta_t)$ is formulated by,
\begin{equation}
\small
    V(s_t+\delta_t) = \mathbb{E}\left[\sum\nolimits_{k=1}^{T}\gamma^k r(s_{t+k}+\delta_t)|\pi_{\theta^S}\right]. 
\end{equation}
Note that $\delta_t$ is only applied to $s_t$, reflecting the impact for state value after adding $\delta_t$ on state $s_t$.

According to both Theorem 1 in \cite{achiam2017constrained} and and Theorem 5 in \cite{zhang2020robust}, the difference between $V(s_t)$ and $V(s_t+\delta_t)$ can be bounded as,
\begin{equation}
\small
    \max_{s_t\in \mathcal{S}}\{V(s_t)-V(s_t+\delta_t)\}\leqslant\alpha\max_{s_t\in \mathcal{S}} \max_{\norm{\delta_t}\leqslant \epsilon}\mathcal{D}(\pi_{\theta^S}(s_t), \pi_{\theta^S}(s_t+\delta_t)),
    \label{theorem5}
\end{equation}
where $\mathcal{D}(\pi_{\theta^S}(s_t), \pi_{\theta^S}(s_t+\delta_t))$ is the distance in action space between the student policy {prescription} $\pi_{\theta^S}(s_t)$ without adversary and the prescription $\pi_{\theta^S}(s_t+\delta_t)$ with adversary;  \begin{small}$\alpha:=2\left[1+\frac{\gamma}{(1-\gamma)^2}\max_{s_t,a_t,s_{t+1}\in\mathcal{S}\times\mathcal{A}\times\mathcal{S}}[r(s_t, a_t)]\right]$\end{small} is a constant independent on $\pi_{\theta^S}$.
Eq.~(\ref{theorem5}) indicates that the \textit{state value gap between $V(s_t)$ and $V(s_t+\delta_t)$ has an upper bound relying on the distance $\mathcal{D}(\pi_{\theta^S}(s_t), \pi_{\theta^S}(s_t+\delta_t))$ in action space}. This motivates us to improve the adversarial robustness of the student policy by directly minimizing $\mathcal{D}(\pi_{\theta^S}(s_t), \pi_{\theta^S}(s_t+\delta_t))$ during policy distillation.

Recalling the discrete action selection in RL, given a deterministic policy $\pi_{\theta^S}$, the action is selected as $\argmax_a \pi_{\theta^S}(s_t,a)$. The distance $\mathcal{D}(\pi_{\theta^S}(s_t), \pi_{\theta^S}(s_t+\delta_t))$ can thus be defined as,
\begin{equation}
\label{eq:distance}
\small
    \mathcal{D}(\pi_{\theta^S}(s_t), \pi_{\theta^S}(s_t+\delta_t)) = \begin{cases}
    0, \text{ } &\argmax_a \pi_{\theta^S}(s_t,a) = \\ &\argmax_a \pi_{\theta^S}(s_t+\delta_t,a)\\ 
1, &\text{  }\text{otherwise}
\end{cases}.
\end{equation}
Note that in policy distillation, the student policy is trained to be consistent with its teacher policy, viz., selecting the same action $a^T$ from $\pi_{\theta^T}$ as
\begin{equation}
    \argmax_a\pi_{\theta^T}(s_t,a) = a^T= \argmax_a \pi_{\theta^S}(s_t,a).
\end{equation} 

Therefore, to minimize the distance in Eq.~(\ref{eq:distance}), $a^T = \argmax_a \pi_{\theta^S}(s_t+\delta_t, a)$ is to be ensured. In other words, we need to encourage the student policy $\pi_{\theta^S}$ to choose the action $a^T$ selected by the teacher policy $\pi_{\theta^T}$, even with the adversary $\delta_t$ existing on state $s_t$. Accordingly, the following proposition is put forth for a robust student policy based on a pre-trained teacher policy $\pi_{\theta^T}$.

\begin{propositions}{\textit{(Robust student policy)}}
We assume that $\pi_{\theta^S}$ and $\pi_{\theta^T}$ are deterministic policies. The optimal action chosen by the teacher policy $\pi_{\theta^T}$ is $a^T =\argmax_a \pi_{\theta^T}(s_t,a)$. Given bounded adversarial perturbations {$\delta_t,\norm{\delta_t}\leqslant \epsilon$} on state $s_t$, we define the prescription gap of student policy $\pi_{\theta^S}$ as,
\begin{equation}
\footnotesize
\begin{split}
    \mathcal{G}_{\theta^S}(s_t+\delta_t, a^T) &= \min_{\delta_t,\norm{\delta_t} \leqslant \epsilon} \left[ \pi_{\theta^S}(s_t+\delta_t, a^T)-\pi_{\theta^S}(s_t+\delta_t, a) \right]\\
    &, \forall a\in \mathcal{A} \cap a\neq a^T.\\
\end{split}
\label{eq:theorem5}
\end{equation}
Then, a robust distilled student policy $\pi_{\theta^S}$ \textbf{must guarantee} $\mathcal{G}_{\theta^S}(s_t+\delta_t, a^T)>0$.
\label{theorem:1}
\end{propositions}
Proposition \ref{theorem:1} indicates that as long as $\mathcal{G}_{\theta^S}(s_t+\delta_t, a^T)> 0$, the action selected by the distilled student policy under any adversarial perturbation $\norm{\delta_t}\leqslant \epsilon$ on state $s_t$ is still $a^T$. Hence, during policy distillation, the value of $\pi_{\theta^S}(s_t+\delta_t, a^T)-\pi_{\theta^S}(s_t+\delta_t, a)$ needs to be maximized in order to boost adversarial robustness. According to Taylor expansion, we have
\begin{equation}
\begin{split}
   \pi_{\theta^S}(s_t+\delta_t, a^T)  &= \pi_{\theta^S}(s_t, a^T) + \delta_t \nabla_{s_t} \pi_{\theta^S}(s_t, a^T) + \omega_1,\\
   \pi_{\theta^S}(s_t+\delta_t, a) &= \pi_{\theta^S}(s_t, a) + \delta_t \nabla_{s_t} \pi_{\theta^S}(s_t, a) + \omega_2,\\
\end{split}
\end{equation}
where $\omega_1$ and $\omega_2$ are truncation errors. For ease of analysis we make a common assumption, viz., $\omega_1-\omega_2 =0$. Thereby, Eq.~(\ref{eq:theorem5}) can be transformed as,
\begin{equation}
\footnotesize
\begin{split}
    \mathcal{G}_{\theta^S}(s_t+\delta_t, a^T)&  = \min_{\delta,\norm{\delta_t}\leqslant \epsilon} \left[(\pi_{\theta^S}(s_t, a^T) + \delta_t \nabla_{s_t}\pi_{\theta^S}(s_t, a^T))- \right. \\
    &\left. (\pi_{\theta^S}(s_t, a) + \delta_t \nabla_{s_t} \pi_{\theta^S}(s_t, a)) \right]\\
     &= \left[\pi_{\theta^S}(s_t, a^T)-\pi_{\theta^S}(s_t, a)\right] + \\
     &\min_{\delta_t,\norm{\delta_t}\leqslant \epsilon}\left[\delta_t \nabla_{s_t}\pi_{\theta^S}(s_t, a^T) -\delta_t \nabla_{s_t}\pi_{\theta^S}(s_t, a)  \right]\\
     &= \left[\pi_{\theta^S}(s_t, a^T)-\pi_{\theta^S}(s_t, a)\right] + \\
     &\min_{\delta_t,\norm{\delta_t}\leqslant \epsilon} \delta_t  \nabla_{s_t}\left[\pi_{\theta^S}(s_t, a^T) -\pi_{\theta^S}(s_t, a)  \right]\\
    & = \mathcal{G}_{\theta^S}(s_t, a^T) + \min_{\delta_t,\norm{\delta_t}\leqslant \epsilon} \delta_t \nabla_{s_t}\mathcal{G}_{\theta^S}(s_t, a^T).
\end{split}
\label{eq:maximize_gap}
\end{equation}

Thus, in order to maximize the prescription gap \begin{small}$\mathcal{G}_{\theta^S}(s_t+\delta_t, a^T)$\end{small}, the first term \begin{small}$\mathcal{G}_{\theta^S}(s_t, a^T)$\end{small} in Eq. (\ref{eq:maximize_gap}) should be maximized.
Note that $\delta_t$ is optimized by an attacker to impact \begin{small}$\mathcal{G}_{\theta^S}(s_t, a^T)$\end{small} \textbf{negatively}; thus being unable to be controlled. Although $\delta_t$ can not be controlled by $\pi_{\theta^S}$, we can alternatively control its impact weight (i.e., the Jacobian \begin{small}$  \nabla_{s_t}\mathcal{G}_{\theta^S}(s_t, a^T)$\end{small}). Namely, to improve the robustness, the influence of the second term in Eq. (\ref{eq:maximize_gap}) can be reduced by minimizing the magnitude of the Jacobian \begin{small}$  \norm{\nabla_{s_t}\mathcal{G}_{\theta^S}(s_t, a^T)}$\end{small}.


%
In sum, to maximize the prescription gap $\mathcal{G}_{\theta^S}(s_t+\delta_t, a^T)$ with attack, we can alternatively maximize
$\mathcal{G}_{\theta^S}(s_t, a^T)$ 
\textbf{without} attack
and simultaneously minimize the magnitude of Jacobian \begin{small}
$\norm{\nabla_{s_t}\mathcal{G}_{\theta^S}(s_t, a^T)}$\end{small}. 
{It is noteworthy that our robustness improvement derived in Eq. (\ref{eq:maximize_gap}) is different from SA-MDP \cite{zhang2020robust}, although both works are motivated by a similar bound relationship as shown in Eq. (\ref{theorem5}). The goal of SA-MDP~\cite{zhang2020robust} is to align the prescriptions with adversary and without adversary; thus querying and generating the perturbation $\delta_t$ is still required. On the contrary, we can ignore the usage of any knowledge from $\delta_t$ by directly regularizing the prescription gap and the corresponding input gradient.}
Guided by this finding, we devise our adversary agnostic policy distillation (A2PD) loss in what follows.

\subsection{Adversary Agnostic Policy Distillation (A2PD)}
\label{section.loss}
To enable the adversary agnostic robust DRL, our policy distillation loss $\mathcal{L}_{A2PD}(\theta^S)$ is proposed as,
\begin{equation}
    \mathcal{L}_{A2PD}(\theta^S) =  \mathcal{L}_{pgm}(\theta^S) + \beta\mathcal{L}_{jr}(\theta^S),
    \label{eq:pd loss}
\end{equation}
where $\mathcal{L}_{pgm}(\theta^S)$ is the {prescription gap maximization loss} that not only maximizes the likelihood on action $a^T$ selected by teacher policy $\pi_{\theta^T}$, but also maximizes the entropy on the remaining actions; $\mathcal{L}_{jr}(\theta^S)$ is the Jacobian regularization loss that aims to boot the robustness via minimizing the magnitude of input gradient on $s_t$ back-propagated from $\mathcal{L}_{pgm}(\theta^S)$. 
Recalling Eq. (\ref{eq:maximize_gap}), minimizing $\mathcal{L}_{pgm}(\theta^S)$ and $\mathcal{L}_{jr}(\theta^S)$ corresponds to maximizing $\mathcal{G}_{\theta^S}(s_t, a^T)$ and minimizing $\nabla_{s_t}\mathcal{G}_{\theta^S}(s_t, a^T)$, respectively. 
The weight $\beta$ controls the strength of $\mathcal{L}_{jr}(\theta^S)$.
We illustrate the details of $\mathcal{L}_{pgm}(\theta^S)$ and $\mathcal{L}_{jr}(\theta^S)$ as follows.

\textbf{Prescription Gap  Maximization {(PGM)}.} With the goal of maximizing the prescription gap between action $a^T$ and the remaining actions in mind, we devise the PGM loss $\mathcal{L}_{pgm}(\theta^S)$ as,
\begin{equation}
\small
\begin{split}
    \mathcal{L}_{pgm}(\theta^S)& = -\pi_{\theta^S}(s_t,a^T)^{\eta}\cdot \left[-  \sum_{a=1,a\neq a^T}^{|\mathcal{A}|}\left(\frac{\pi_{\theta^S}(s_t,a)}{1-\pi_{\theta^S}(s_t,a^T)}\right) \right.\\
    &\left. \log\left(\frac{\pi_{\theta^S}(s_t,a)}{1-\pi_{\theta^S}(s_t,a^T)}\right)\right],\\
\end{split}
\label{eq:pgm}
\end{equation}
where $\pi_{\theta^S}(s_t,a)$ is the prescription on action $a$; $\eta\in(0,1)$ is a constant.
The rationale behind Eq.~(\ref{eq:pgm}) is that {minimizing} $\mathcal{L}_{pgm}(\theta^S)$ enables to simultaneously maximize the likelihood $\pi_{\theta^S}(s_t, a^T)$ on the action $a^T$ (i.e., the action selected by the teacher policy $\pi_{\theta^T}$) and the entropy over the remaining actions 
$-\sum_{a=1,a\neq a^T}^{\vert\mathcal{A}\vert}(\frac{\pi_{\theta^S}(s_t,a)}{1-\pi_{\theta^S}(s_t,a^T)})\log(\frac{\pi_{\theta^S}(s_t,a)}{1-\pi_{\theta^S}(s_t,a^T)})$. The entropy maximization results in a smaller maximum over action $a, \text{ }a\in\mathcal{A},\text{}a\neq a^T$. 
Hence, by maximizing $\pi_{\theta^S}(s_t, a^T)$ at the same time, we can facilitate a larger prescription gap $\mathcal{G}_{\theta^S}(s_t,a^T)$.
Note that, the entropy calculation is weighted by $\frac{1}{1-\pi_{\theta^S}(s_t, a^T)}$; this makes the distillation training focus on maximizing $\pi_{\theta^S}(s_t, a^T)$ at the beginning when $\pi_{\theta^S}(s_t, a^T)$ is small. In contrast, when $\pi_{\theta^S}(s_t, a^T)$ increases during training, $\mathcal{L}_{pgm}(\theta^S)$ gradually shifts attention to the entropy maximization.
In addition, $\eta$ balances the maximization on  $\pi_{\theta^S}(s_t, a^T)$ and entropy regularization, which is analyzed in Section 4.5.


\textbf{Jacobian Regularization (JR).} As derived in Eq.~(\ref{eq:maximize_gap}), a robust policy distillation also requires minimizing the Jacobian on the input $s_t$ as additional regularization. The concept of JR was introduced by Drucker and Le Cun \cite{drucker1992improving} in double backpropagation to enhance generalization performance, where they trained neural networks not only by minimizing the gradient on weights but the gradient with respect to the input features. Hoffman et al. \cite{hoffman2019robust} and Chan et al. \cite{chan2019jacobian} utilized JR to regularize the stability and interpretability of image classifiers. However, how to effectively exploit JR in RL adversarial robustness, especially in the policy distillation process, has so far remained under-explored.
With that in mind, we thus propose the JR loss,
\begin{equation}
\footnotesize
    \mathcal{L}_{jr}(\theta^S) =   \norm{\frac{\partial \mathcal{L}_{pgm}(\theta^S)}{\partial s_t}}_F,
    \label{eq:jr_loss}
\end{equation}
where $\frac{\partial \mathcal{L}_{pgm}(\theta^S)}{\partial s_t}$ {indicates the Jacobian on state $s_t$ w.r.t. the loss function} $\mathcal{L}_{pgm}(\theta^S)$; $F$ represents the Frobenius norm.
It is worth noting that most start-of-the-art attack algorithms (e.g., FGSM \cite{goodfellow2015explaining}) are on the basis of utilizing the Jacobian, thus minimizing the magnitude of Jocobian intuitively provides weaker gradient information; this makes a harder generation of $\delta_t$ for an attacker. 
In addition, according to the analysis in~\cite{finlay2019scaleable}, if we maximize the prescription gap, it is able to alleviate the issue of gradient masking. A more detailed analysis on the improvement of adversarial robustness via minimizing the magnitude of Jacobian is provided in Theorem \ref{theorem:3}.

\subsection{Theoretical Analysis}
To support the our loss design for robust policy distillation, we analyze the policy prescription gap and the resultant improvement on adversarial robustness in the following theoretical analysis. 

\begin{theorem}{\textit{(Policy prescription gap maximization)}}
Given a particular prescription $\pi_{\theta^S}(s_t, a^T)$ by student policy $\pi_{\theta^S}$ on the action $a^T$, if the PGM loss $\mathcal{L}_{pgm}(\theta^S)$ is minimized, it is guaranteed that the prescription gap $\mathcal{G}_{\theta^S}(s_t, a^T)$ in Eq.~(\ref{eq:maximize_gap}) is maximized. Moreover, if $\pi_{\theta^S}(s_t, a^T)>\frac{1}{|\mathcal{A}|}$ where $|\mathcal{A}|$ is the size of action space, $\mathcal{G}_{\theta^S}(s_t, a^T)$ is ensured to be positive. 
\label{theorem:2}
\end{theorem}

The proof follows from the fact that maximum entropy is attained when the distribution over actions is uniform, which is shown as below.

\begin{proof}
Given the condition that $\pi_{\theta^S}(s_t,a^T)$ is a particular prescription, we define $\pi_{\theta^S}(s_t,a^T) = C$ as a constant. Then the PGM loss in Eq.~(\ref{eq:pgm}) can be rewritten as,
\begin{equation}
\scriptsize
    \mathcal{L}_{pgm}(\theta^S) = -C^{\eta} \cdot \left[-\sum_{a=1,a\neq a^T}^{|\mathcal{A}|}\left(\frac{\pi_{\theta^S}(s_t,a)}{1-C}\right)\log\left(\frac{\pi_{\theta^S}(s_t,a)}{1-C}\right)\right],
\end{equation}
As $C\in (0,1)$, we can get $1-C\in (0,1)$ and $C^\eta\in (0,1)$ where $\eta$ is a positive constant. Therefore, minimizing $\mathcal{L}_{pgm}(\theta^S)$ is equal to maximizing the entropy $h(\pi_{\theta^S}(s_t,a))$,
\begin{equation}
\small
     h(\pi_{\theta^S}(s_t,a)) = -\sum_{a=1,a\neq a^T}^{|\mathcal{A}|}\pi_{\theta^S}(s_t,a)\log(\pi_{\theta^S}(s_t,a)).
\end{equation}
Given \begin{small}$\pi_{\theta^S}(s_t,a^T) = C$\end{small}, we have  \begin{small}$\sum_{a=1,a\neq a^T}\pi_{\theta^S}(s_t,a) = 1-C $\end{small}. According to the information theorem~\cite{gray2011entropy}, the maximum of $h(\pi_{\theta^S}(s_t,a))$ is obtained  when the distribution of $\pi_{\theta^S}(s_t,a)$ is uniform; this results in a minimum \begin{small}$\pi_{\theta^S}(s_t,a) = \frac{1-C}{|\mathcal{A}|-1}$\end{small}. Thereby, we can get the maximized prescription gap
\begin{equation}
\small
\begin{split}
    \mathcal{G}_{\theta^S}(s_t,a^T)& = C-\frac{1-C}{|\mathcal{A}|-1}\\
    & = \frac{|\mathcal{A}|}{|\mathcal{A}|-1}C-\frac{1}{|\mathcal{A}|-1}.
\end{split}
\end{equation}
If $\pi_{\theta^S}(s_t,a^T)=C>\frac{1}{|\mathcal{A}|}$, then we can get $\mathcal{G}_{\theta^S}(s_t,a^T)>0$.
\end{proof}

\begin{theorem}{\textit{(Adversarial robustness)}}
Given a student policy $\pi_{\theta^S}$, if both the PGM loss $\mathcal{L}_{pgm}(\theta^S)$ and JR loss $\mathcal{L}_{jr}(\theta^S)$ are minimized, we can guarantee an improvement on adversarial robustness.
\label{theorem:3}
\end{theorem}

The basic idea for proof is that the minimized \begin{small}$\norm{\frac{\partial \mathcal{L}_{pgm}(\theta^S)}{\partial s_t}}$\end{small} minimizes the impact of adversary $\delta_t$ on the PGM loss. The detail of proof is shown as below.
\begin{proof}
In our policy distillation, the PGM loss $\mathcal{L}_{pgm}(\theta^S)$ is optimized to simultaneously maximize the probability on action $a^T$ and the prescription gap $\mathcal{G}_{\theta^S}(s_t,a^T)$ as shown in the proof of Theorem \ref{theorem:2}. Therefore, 
in order to keep the policy $\pi_{\theta^S}$ robust to adversary $\delta_t, \norm{\delta_t}\leqslant\epsilon$, we have to ensure the PGM loss is still minimized with adversary $\delta_t$. For ease of analysis, we rewrite the PGM loss $\mathcal{L}_{pgm}(\theta^S)$ as a function of $s_t$,
\begin{equation}
\footnotesize
\begin{split}
    \mathcal{L}_{pgm}(s_t) &= -\pi_{\theta^S}(s_t,a^T)^{\eta} \cdot \left[-\sum_{a=1,a\neq a^T}^{|\mathcal{A}|}\left(\frac{\pi_{\theta^S}(s_t,a)}{1-\pi_{\theta^S}(s_t,a^T)}\right) \right. \\
   & \left. \log \left(\frac{\pi_{\theta^S}(s_t,a)}{1-\pi_{\theta^S}(s_t,a^T)}\right)\right].
\end{split}
\end{equation}
The PGM loss under adversary $\delta_t$ is $\mathcal{L}_{pgm}(s_t+\delta_t)$. Using Taylor expansion, we have
\begin{equation}
\begin{split}
     \mathcal{L}_{pgm}(s_t+\delta_t) &= \mathcal{L}_{pgm}(s_t) + \delta_t \frac{\partial\mathcal{L}_{pgm}(s_t)}{s_t} +\omega_{pgm},\\ &\forall \delta_t, \norm{\delta_t}\leqslant\epsilon.
\end{split}
\end{equation}
$\omega_{pgm}$ is the truncation error in Taylor expansion, thus it can be assumed small enough to be ignored.
Note that $\delta_t$ is generated by the attacker to negatively impact our distilled policy, which makes $\delta_t$ not controllable from the distillation training perspective. Therefore, to minimize the impact of $\delta_t \frac{\partial\mathcal{L}_{pgm}(s_t)}{s_t}$, we can alternatively minimize the magnitude of Jacobian \begin{small}$\norm{\frac{\partial\mathcal{L}_{pgm}(s_t)}{s_t}}$\end{small}. 

Hence, given that the $\mathcal{L}_{pgm}(s_t)$ \textbf{without} adversary is minimized, if we minimize the norm of Jacobian \begin{small}$\norm{\frac{\partial\mathcal{L}_{pgm}(s_t)}{s_t}}$\end{small}, we can still achieve a minimized PGM loss $\mathcal{L}_{pgm}(s_t+\delta_t)$ \textbf{with} adversary $\delta_t$. Therefore, even with adversarail atttack $\delta_t$, the loss $\mathcal{L}_{pgm}(s_t+\delta_t)$ can still be minimized, which leads to a maximized $\mathcal{G}_{\theta^S}(s_t,a^T)>0$ as proved in Theorem~\ref{theorem:2}. Accordingly, we can prove that the adversarial robustness is improved.
\end{proof}

\section{Experiments}
\subsection{Experimental Setup}
\begin{figure}[t]
    \centering
       \begin{subfigure}{0.2\linewidth}
        \centering
        \includegraphics[width=0.8\textwidth]{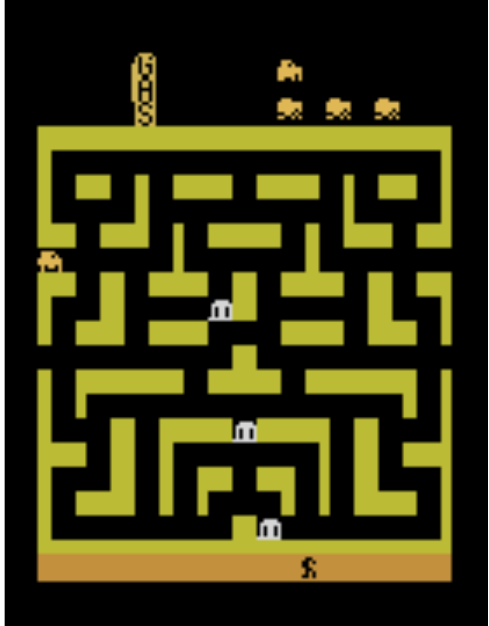}
        \caption{BankHeist}
    \end{subfigure}%
    \begin{subfigure}{0.2\linewidth}
        \centering
        \includegraphics[width=0.8\textwidth]{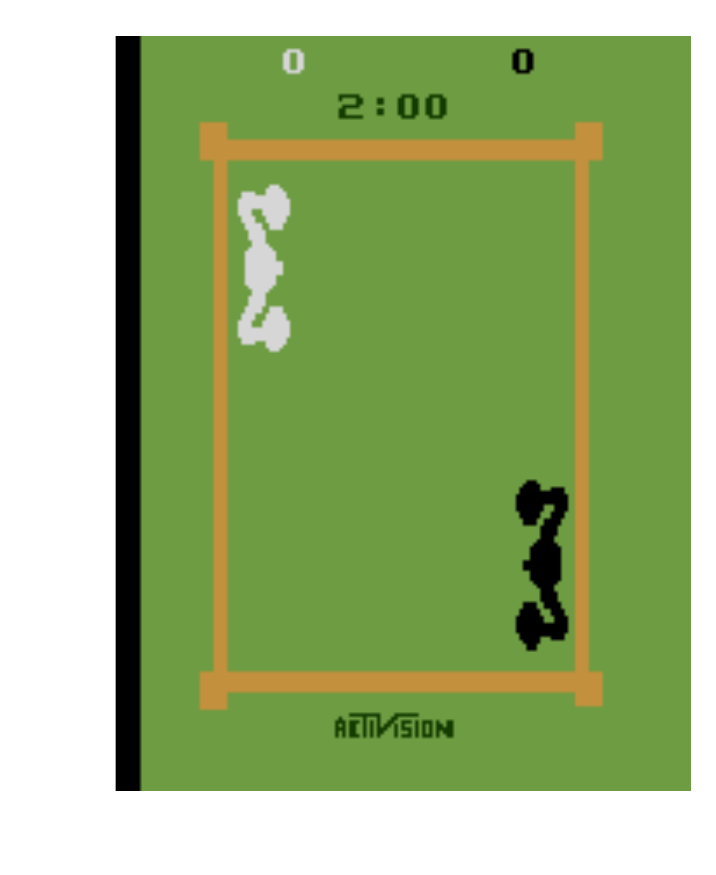}
        \caption{Boxing}
    \end{subfigure}%
    \begin{subfigure}{0.2\linewidth}
        \centering
        \includegraphics[width=0.8\textwidth]{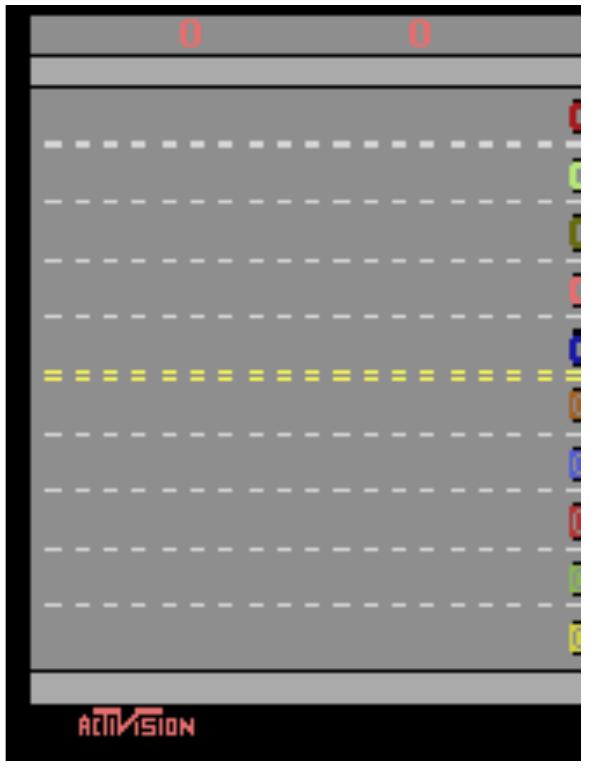}
        \caption{Freeway}
    \end{subfigure}%
    \begin{subfigure}{0.2\linewidth}
        \centering
        \includegraphics[width=0.8\textwidth]{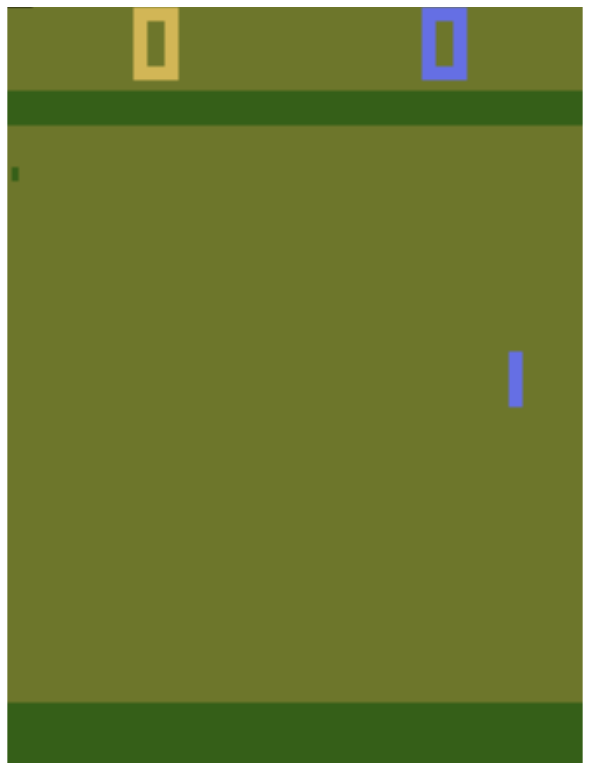}
        \caption{Pong}
    \end{subfigure}%
    \begin{subfigure}{0.2\linewidth}
        \centering
        \includegraphics[width=0.8\textwidth]{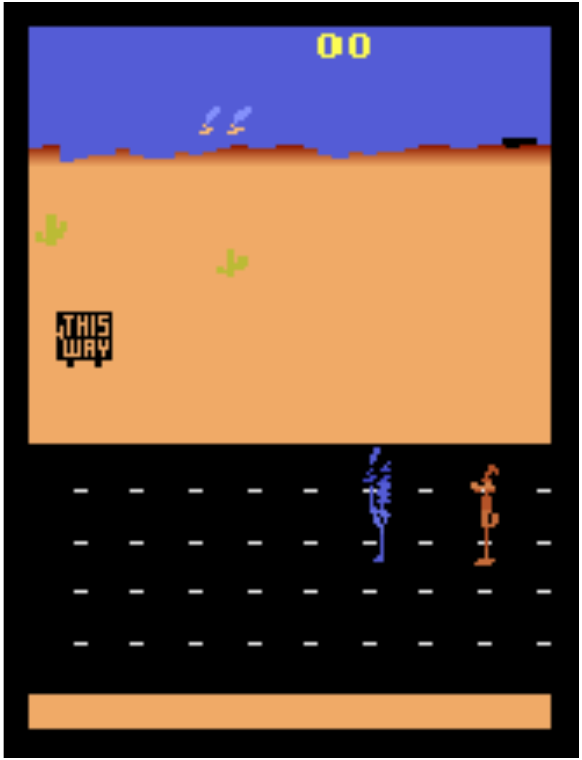}
        \caption{RoadRunner}
    \end{subfigure}%
    \caption{The visualization of the initial state $\mathbf{s}_t$ of the five selected Atari games.}
    \label{fig:game screen display}
    \vspace{-0.2in}
\end{figure}

We test our approach\footnote{Our code will be released upon acceptance.} on five Atari games (i.e., Freeway, Bank Heist, Pong, Boxing and Road Runner) that are utilized in the state-of-the-arts \cite{fischer2019online,zhang2020robust}. The visualization of each game screen is shown in Fig. \ref{fig:game screen display}. For each game, the state is a 4-stack of consecutive frames, where each frame is {pre-processed} to size $84\times 84$. The pre-processing applies the environment wrapper (based on Arcade learning environment) in Rainbow. After preprocessing, the pixel value from $[0,255]$ is normalized to $[0,1]$. 

\noindent\textbf{Teacher Policy Training.} In our evaluation, the teacher policy is trained by RAINBOW-DQN~\cite{hessel2018rainbow} that is deemed to be the start-of-the-art with combining many ingradients, such as distributional RL \cite{bellemare2017distributional}, dueling networks \cite{wang2016dueling} and noisy nets \cite{fortunato2018noisy}. Each teacher policy is trained with 4 million frames on a particular game (see source code\footnote{https://github.com/Kaixhin/Rainbow}), which costs 12-40 hours on Nvidia 2080Ti. The other parameter settings for teacher policy training are provided in Table~\ref{tab:parameter setting} in Appendix. More detailed explanations of each parameter can refer to~\cite{hessel2018rainbow}.

\noindent\textbf{Student Policy Distillation Training.} In our implementation, the network structure of student policy $\pi_{\theta^S}$ uses the Nature-DQN structure \cite{mnih2015human}. To train the student policy, we collect $1\times 10^5$ state-prescription pairs $[s_t, \pi_{\theta^T}(s_t)]$ from the teacher policy $\theta^T$, where $90\%$ of the collected data is treated as training and the remaining $10\%$ as testing. We use Adam as the optimizer, and the implementation is based on Keras. The rest hyperparameter settings for policy distillation are provided in Table~\ref{tab:student_parameter_setting} in Appendix.

\noindent\textbf{Evaluation of Robustness.} 
To align with the compared state-of-the-arts \cite{fischer2019online,zhang2020robust,behzadan2017whatever}, 
we use the untargeted Projected Gradient Decent (PGD) attack. However, to evaluate our achieved robustness more generally, we further tested our distilled policies using two additional attacks, including Jacobian saliency map attack (JSMA) \cite{papernot2016limitations} and fast gradient sign method (FGSM) \cite{goodfellow2015explaining}. The PGD attack performs $K$-iteration updates of adversary $\delta_t$, given by:
\begin{equation}
\small
\begin{split}
    s^{k+1} &= s^{k} + \frac{\epsilon}{K} \mathcal{P}\left(\frac{\partial\mathcal{H}(\pi_{\theta^S}(s^k),a^T)}{\partial s^k}\right), \\
    &\text{  }s^0 =s_t, \text{  }k=0,1,\cdots, K-1,\\
\end{split}
\label{eq:pgd attack}
\end{equation}
where $s^{k+1}$ is the attacked input state with adversarial perturbation inside; $\mathcal{H}(\pi_{\theta^S}(s^k),a^T)$ is the cross-entropy loss between student policy prescription $\pi_{\theta^S}(s^k)$ and the one-hot vector encoded based on action $a^T$ selected by teacher policy. $\mathcal{P}$ is an operator for projecting the input gradient into a constrained norm ball. $\epsilon$ and
$\frac{\epsilon}{K}$ are the total norm constraint and the norm constraint for each iteration step, respectively.  
We explore three different values of $K\in\{4, 10, 50\}$, and two different values of $\epsilon\in\{1/255, 3/255\}$. Note that $\epsilon=3/255$ is a stronger attack that has not been evaluated in previous studies \cite{fischer2019online,zhang2020robust}. 
The implementation of all the evaluated attacks is based on the adversarial robustness toolbox \cite{art2018}. 

\begingroup
\begin{table*}[t]
\centering   
\scriptsize
    \renewcommand{\arraystretch}{1}
	\caption{The comparison of averaged accumulated reward on five start-of-the-art methods and our adversary agnostic PD (A2PD). In evaluating A2PD, the averaged accumulated rewards is over 50 episodes. Note that, NA in table represents the evaluation is not reported in previous studies. Natural reward indicates the reward obtained without any attack, which corresponds to the $\mathcal{R}_B$ in Eq.~(\ref{eq:relative_robustness}). The results of vanilla DQN and \textit{DQN Adv. Training} \cite{behzadan2017whatever} are imported from \cite{zhang2020robust}.}
	\addtolength{\tabcolsep}{-3pt}
    \begin{tabular}{|c|c|c|c|c|c|c|}
    \specialrule{.18em}{.05em}{.05em}
    Approaches &Attack Setting & Bank Heist & Boxing & Freeway & Pong & Road Runner\\
    \specialrule{.1em}{.05em}{.05em}
    \multirow{3}{*}{\makecell{DQN \\ (Vanilla)}} 
    &Natural Reward & $1308.4.0\pm 24.1$  & $70.9\pm34.1$ &$34.0\pm 0.2$ &$21.0\pm 0.0$ &$45534.0\pm7066.0$\\ 
    \arrayrulecolor{black!50}\cline{2-7}
    & \cellcolor{blue!10}\makecell{PGD Attack Reward\\ ($K=10,\epsilon=\frac{1}{255}$)} &\cellcolor{blue!10} $564.0\pm 21.2$ & \cellcolor{blue!10}$4.8\pm5.5$ &\cellcolor{blue!10} $0.0\pm 0.0$ &\cellcolor{blue!10} $-21.0 \pm 0.0$ & \cellcolor{blue!10}$0.0\pm0.0$\\ 
    \specialrule{.15em}{.05em}{.05em}
    \multirow{3}{*}{\makecell{DQN Adv. Training \cite{behzadan2017whatever}\\ (attack 50\% frames)}} 
    &Natural Reward & $1126.0\pm 70.9$ & NA &$25.4\pm 0.8$ & $10.1\pm 6.6$ &$22944.0\pm6532.5$\\ 
    \arrayrulecolor{black!50}\cline{2-7}
    &\cellcolor{blue!10}\makecell{PGD Attack Reward\\ ($K=10,\epsilon=\frac{1}{255}$)} &\cellcolor{blue!10} $9.4\pm 13.6$ &\cellcolor{blue!10} NA &\cellcolor{blue!10} $0.0\pm 0.0$ & \cellcolor{blue!10}$-21.0 \pm 0.0$ & \cellcolor{blue!10}$14.0\pm34.7$\\ 
    \specialrule{.15em}{.05em}{.05em}
    \multirow{3}{*}{\makecell{Imitation Learning \cite{fischer2019online}\\ (PGD Adv. Training)}}
    &Natural Reward & $238.66$ & $80.67$ &$32.93$ & $19.73$&$12106.67$\\ 
    \arrayrulecolor{black!50}\cline{2-7}
    &\makecell{PGD Attack Reward\\ ($K=4,\epsilon=\frac{1}{255}$)} & $18.13$ & $50.87$ & $32.53$ &$\textbf{19.67}$&$5753.33$\\ 
    \specialrule{.15em}{.05em}{.05em}
    \multirow{3}{*}{\makecell{SA-DQN \cite{zhang2020robust} \\ (PGD)}} 
    &Natural Reward & $1245.2\pm 14.5$ & NA &$33.9\pm 0.4$ & $21.0\pm 0.0$ &$34032.0\pm3845.0$\\ 
    \arrayrulecolor{black!50}\cline{2-7}
    &\cellcolor{blue!10}\makecell{PGD Attack Reward\\ ($K=10,\epsilon=\frac{1}{255}$)} & \cellcolor{blue!10}$1006.0\pm 226.4$ &\cellcolor{blue!10} NA &\cellcolor{blue!10} $23.7\pm 2.3$ &\cellcolor{blue!10}$21.0 \pm 0.0$ &\cellcolor{blue!10}$20402.0\pm7551.1$\\
    \specialrule{.15em}{.05em}{.05em}
    \multirow{5}{*}{\makecell{SA-DQN \cite{zhang2020robust} \\ (Convex)}} 
    &Natural Reward & $1235.4\pm 9.8$ & NA &$30.0\pm 0.4$ & $21.0\pm 0.0$ &$44638.0\pm7367.0$\\ 
    \arrayrulecolor{black!50}\cline{2-7}
    &\cellcolor{blue!10}\makecell{PGD Attack Reward\\ ($K=10,\epsilon=\frac{1}{255}$)} & \cellcolor{blue!10}$1232.4\pm 16.2$ &\cellcolor{blue!10} NA &\cellcolor{blue!10} $30.0\pm 0.0$ &\cellcolor{blue!10}$21.0 \pm 0.0$ &\cellcolor{blue!10}$\color{blue}\textbf{44732.0}\pm\textbf{8059.5}$\\
    \arrayrulecolor{black!50}\cline{2-7}
    &\cellcolor{brown!10}\makecell{PGD Attack Reward\\ ($K=50,\epsilon=\frac{1}{255}$)} & \cellcolor{brown!10}$1234.6\pm 16.6$ &\cellcolor{brown!10} NA &\cellcolor{brown!10} $30.0\pm 0.0$ &\cellcolor{brown!10}$21.0 \pm 0.0$ &\cellcolor{brown!10}$\color{blue}44678.0\pm6954.0$\\
    \specialrule{.15em}{.05em}{.05em}
    \multirow{11}{*}{\makecell{A2PD \\ (\textit{\textbf{Adversary Agnostic}})}} 
    &Natural Reward & $1617.4\pm 34.8$ & $74.2\pm21.2$ &$33.9\pm 0.3$ & $20.7\pm 0.1$ &$29252.0\pm6610.7$\\ 
    \arrayrulecolor{black!50}\cline{2-7}
    &\cellcolor{blue!10}\makecell{PGD Attack Reward\\ ($K=10,\epsilon=\frac{1}{255}$)} 
    & \cellcolor{blue!10}$\textbf{1620.8}\pm \textbf{22.8}$ 
    & \cellcolor{blue!10}$\color{blue}81.9\pm16.9$ &\cellcolor{blue!10} $\color{blue}\textbf{34.0}\pm \textbf{0.0}$ &\cellcolor{blue!10}$\color{blue}\textbf{21.0} \pm \textbf{0.0}$ &\cellcolor{blue!10}$32076.0\pm7910.7$\\
    \arrayrulecolor{black!50}\cline{2-7}
    &\cellcolor{brown!10}\makecell{PGD Attack Reward\\ ($K=50,\epsilon=\frac{1}{255}$)} &\cellcolor{brown!10} $1620.0\pm 40.6$ &\cellcolor{brown!10} $\color{blue}\textbf{85.2}\pm\textbf{3.7}$ &\cellcolor{brown!10} $\color{blue}33.9\pm 0.4$ &\cellcolor{brown!10}$\color{blue}\textbf{21.0} \pm\textbf{0.0}$ &\cellcolor{brown!10}$31078.0\pm6848.3$\\
    \arrayrulecolor{black!50}\cline{2-7}
    &\makecell{PGD Attack Reward\\ ($K=10,\epsilon=\frac{3}{255}$)} & $\color{blue}1622.0\pm 35.0$ & $\textbf{79.0}\pm\textbf{16.2}$ & $33.5\pm 0.9$ &$18.2 \pm 3.7$ &$\textbf{32396.0}\pm\textbf{5623.0}$\\
    \arrayrulecolor{black!50}\cline{2-7}
    &\makecell{PGD Attack Reward\\ ($K=50,\epsilon=\frac{3}{255}$)} & $1606.2\pm 53.9$ & $77.6\pm11.0$ & $32.8\pm 2.1$ &$17.1 \pm 4.0$ &$30622.0\pm7275.7$\\
    \arrayrulecolor{black!50}\cline{2-7}
    & FGSM Attack Reward & $\color{blue}\textbf{1624.8}\pm \textbf{30.8}$ & $78.4\pm22.85$ & $\textbf{33.8}\pm \textbf{0.3}$ &$\color{blue}20.3 \pm 0.3$ &$31324.0\pm5096.3$\\
    \arrayrulecolor{black!50}\cline{2-7}
    & JSMA Attack Reward& $1615.4\pm 43.6$ & $74.5\pm20.9$ & $\color{blue}33.9\pm 0.2$ &$18.7 \pm 2.3$ &$30084.0\pm6886.3$\\
\arrayrulecolor{black}\specialrule{.18em}{.05em}{.05em}
\end{tabular}
\label{table:robustness comparison}
\vspace{-0.12in}
\end{table*}
\endgroup

\noindent\textbf{Evaluation of Relative Robustness.} 
In previous evaluations of adversarial robustness of DRL, the accumulated rewards under attacks (i.e., usually PGD attacks) are treated as the evaluation metric. However, in evaluating the reward under attack, the performance of baseline model has a significant impact; this inspires us to further design an auxiliary evaluation metric named as relative robustness $\mathcal{M}(\mathcal{R}_\delta,\mathcal{R}_{w/o\ \delta})$, 
\begin{equation}
\small
    \mathcal{M}(\mathcal{R}_\delta,\mathcal{R}_{w/o\ \delta}) = \frac{\mathcal{R}_\delta}{\mathcal{R}_{w/o\ \delta}}\cdot 100\%
    \label{eq:relative_robustness}
\end{equation}
where $\mathcal{R}_\delta$ and $\mathcal{R}_{w/o\ \delta}$ are the reward with and without attack, respectively. 
Hence, $\mathcal{M}(\mathcal{R}_\delta,\mathcal{R}_{w/o\ \delta})$ provides a percentage variation of accumulated rewards with respect to the baseline performance $\mathcal{R}_{w/o\ \delta}$, which should be considered when comparisons are based on different baseline policies. As another contribution of this paper, we contend that $\mathcal{M}(\mathcal{R}_\delta,\mathcal{R}_{w/o\ \delta})$ should be treated as a complement metric for evaluating adversarial robustness.

\subsection{Robustness Evaluation of Policy $\pi_{\theta^S}$}
We evaluate the robustness {of} the policies trained by our adversary agnostic PD (A2PD), and compare our results with the state-of-the-art methods, including (1) vanilla DQN, (2) DQN adversarial training \cite{behzadan2017whatever}, (3) imitation learning with adversarial training \cite{fischer2019online}, (4) SA-DQN with PGD \cite{zhang2020robust}, and (5) SA-DQN with convex relaxation \cite{zhang2020robust}.
To provide a comprehensive comparison on the adversarial robustness, we evaluate our distilled policies based on a wide range of attacks, including PGD attacks (with $K\in\{10,50\}, \epsilon\in\{1/255, 3/255\}$), fast gradient sign method (FGSM) and Jacobian saliency map attack (JSMA).
The results are presented in Table \ref{table:robustness comparison}, where the performance under same attack is shaded with same color. For instance, the performance under PGD attack with $K=10, \epsilon=1/255$ and PGD attack with $K=50, \epsilon=1/255$ are shaded by \colorbox{blue!10}{light blue} and \colorbox{brown!10}{light brown}, respectively. Note that the top 1-3 rewards under attacks are highlighted by \textbf{\color{blue}bold blue}, {\color{blue}blue} and \textbf{bold}, respectively.

\begin{table*}[t]
    \centering
    \scriptsize
    \renewcommand{\arraystretch}{1.3}
    \caption{The comparison of relative robustness. The relative robustness value $\mathcal{M}(\mathcal{R}_\delta,\mathcal{R}_{w/o\ \delta})$ is calculated based on Eq. (\ref{eq:relative_robustness}), where $\mathcal{R}_\delta$ is the reward under PGD ($K=10, \epsilon=1/255$) that is shaded with \colorbox{blue!10}{light blue} in Table \ref{table:robustness comparison}, $\mathcal{R}_{w/o\ \delta}$ is the corresponding natural reward without attack. The largest value of $\mathcal{M}(\mathcal{R}_\delta,\mathcal{R}_{w/o\ \delta})$ is highlighted with \textbf{\color{blue}bold blue} text.}
    \begin{tabular}{|c|l|l|l|l|l|}
    \specialrule{.15em}{.05em}{.05em}
    Approach & BankHeist & Boxing & Freeway & Pong & Road Runner\\
    \specialrule{.1em}{.05em}{.05em}
    DQN (Vanilla) & $43.1\%$ & $6.8\%$ & $0\%$ & $-100\%$ & $0\%$ \\
    \specialrule{.05em}{.05em}{.05em}
    DQN  (Adv Training)\cite{behzadan2017whatever} & 0.8\% & NA & 0\% & -207.9\% &0.06\% \\
    \specialrule{.05em}{.05em}{.05em}
    Imitation Learning \cite{fischer2019online} (PGD Adv Training) & 7.6\% & 63.1\% & 98.8\% & 99.7\% &47.5\% \\
    \specialrule{.05em}{.05em}{.05em}
    SA-DQN \cite{zhang2020robust} (PGD) & 80.8\% & NA & 69.9\% & \textbf{\color{blue}100\%} &59.9\% \\
    \specialrule{.05em}{.05em}{.05em}
    SA-DQN (Convex) \cite{zhang2020robust}& 99.8\% & NA & 100\% & \textbf{\color{blue}100\%} &100.2\% \\
    \specialrule{.05em}{.05em}{.05em}
    \textbf{A2PD (Adversary Agnostic)}& \textbf{\color{blue}100.2\%} & \textbf{\color{blue}110.4\%} & \textbf{\color{blue}100.3\%} & \textbf{\color{blue}100\%} &\textbf{\color{blue}109.7\% }\\
    \specialrule{.15em}{.05em}{.05em}
    \end{tabular}
    \label{tab:relative_robustness}
\end{table*}

\begin{figure*}[t]
    \centering
    \begin{subfigure}{0.33\linewidth}
        \centering
        \includegraphics[width=0.8\textwidth]{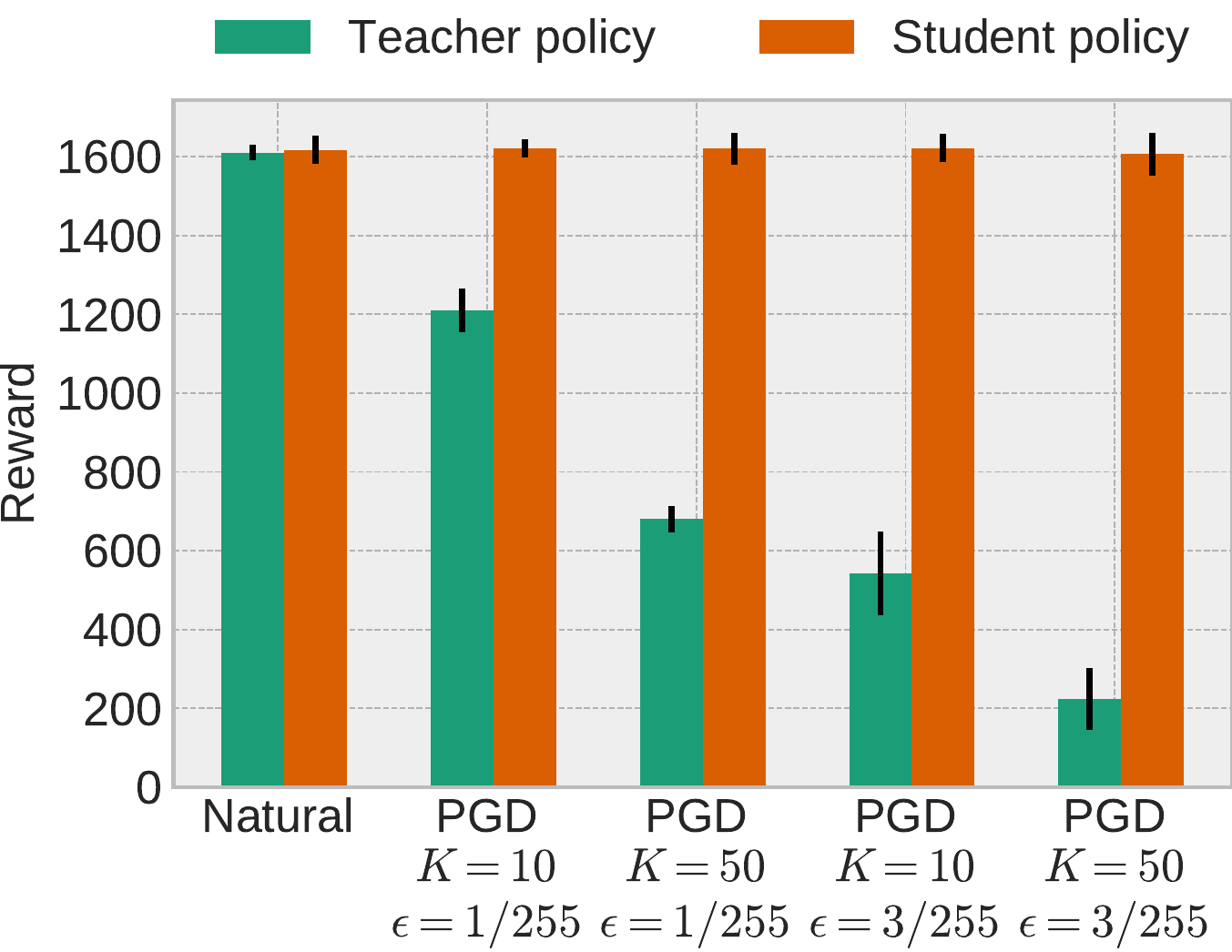}
        \caption{Bank Heist}
    \end{subfigure}%
    \begin{subfigure}{0.33\linewidth}
        \centering
        \includegraphics[width=0.8\textwidth]{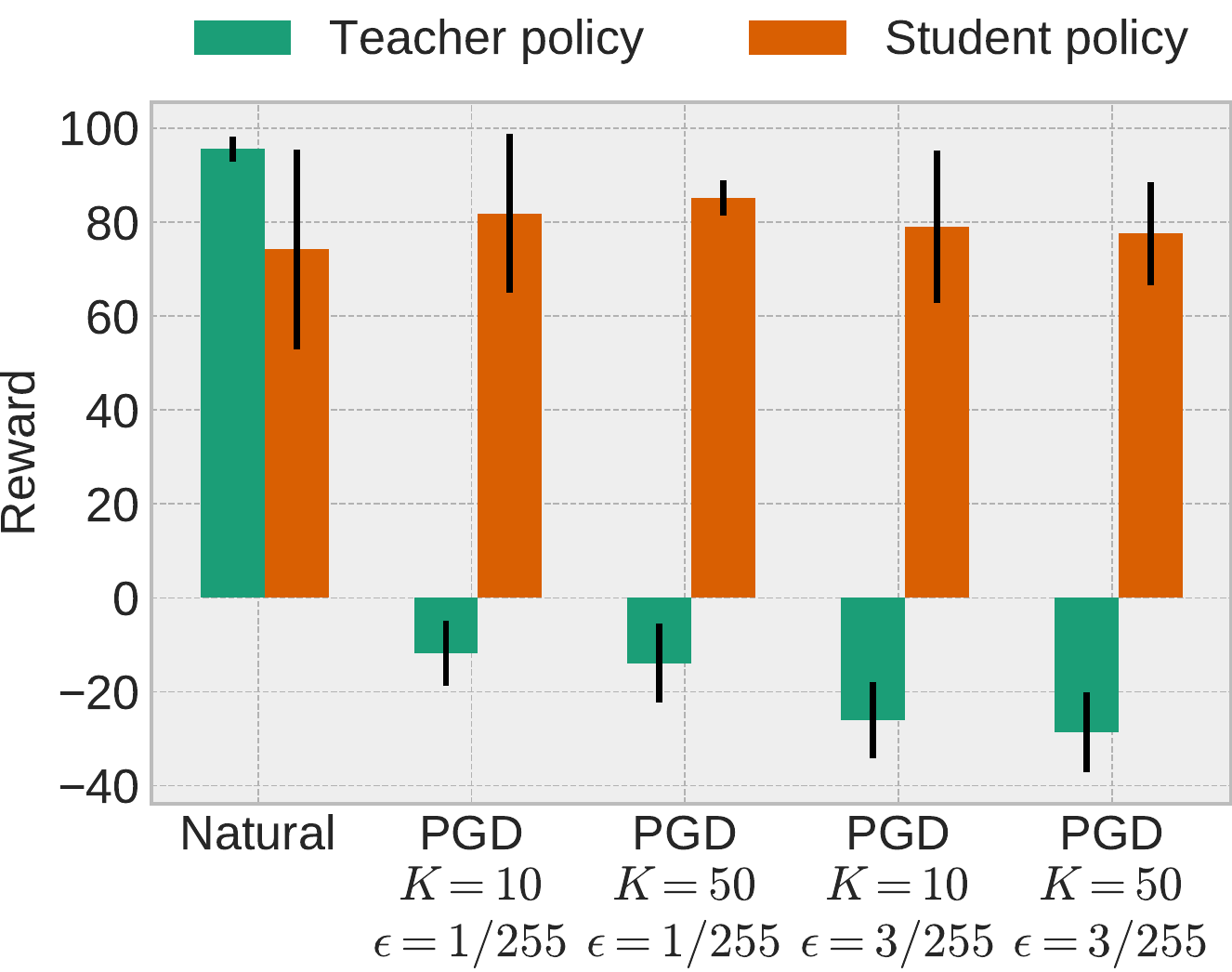}
        \caption{Boxing}
    \end{subfigure}%
    \begin{subfigure}{0.33\linewidth}
        \centering
        \includegraphics[width=0.8\textwidth]{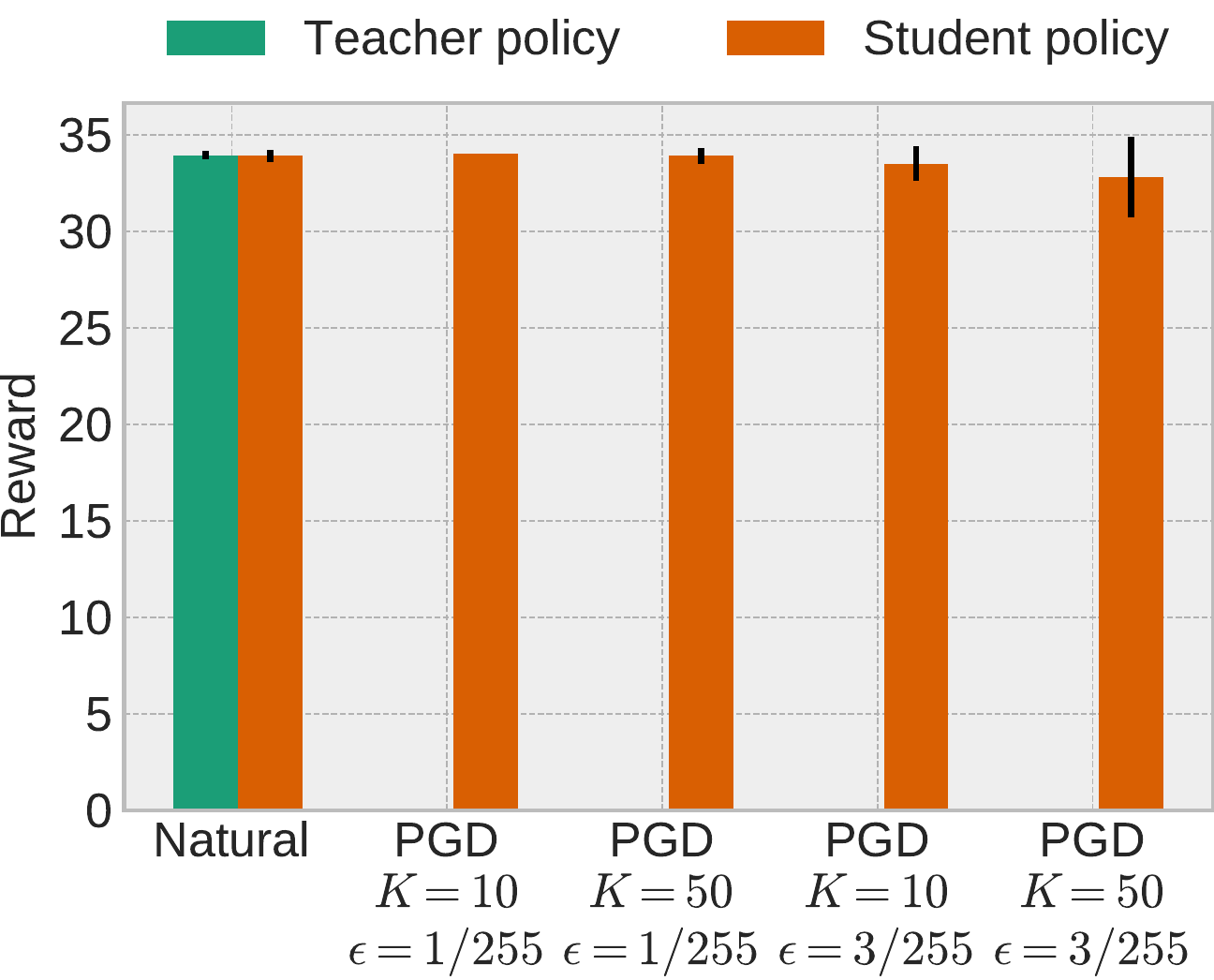}
        \caption{Freeway}
    \end{subfigure}%
    \\
    \begin{subfigure}{0.33\linewidth}
        \centering
        \includegraphics[width=0.8\textwidth]{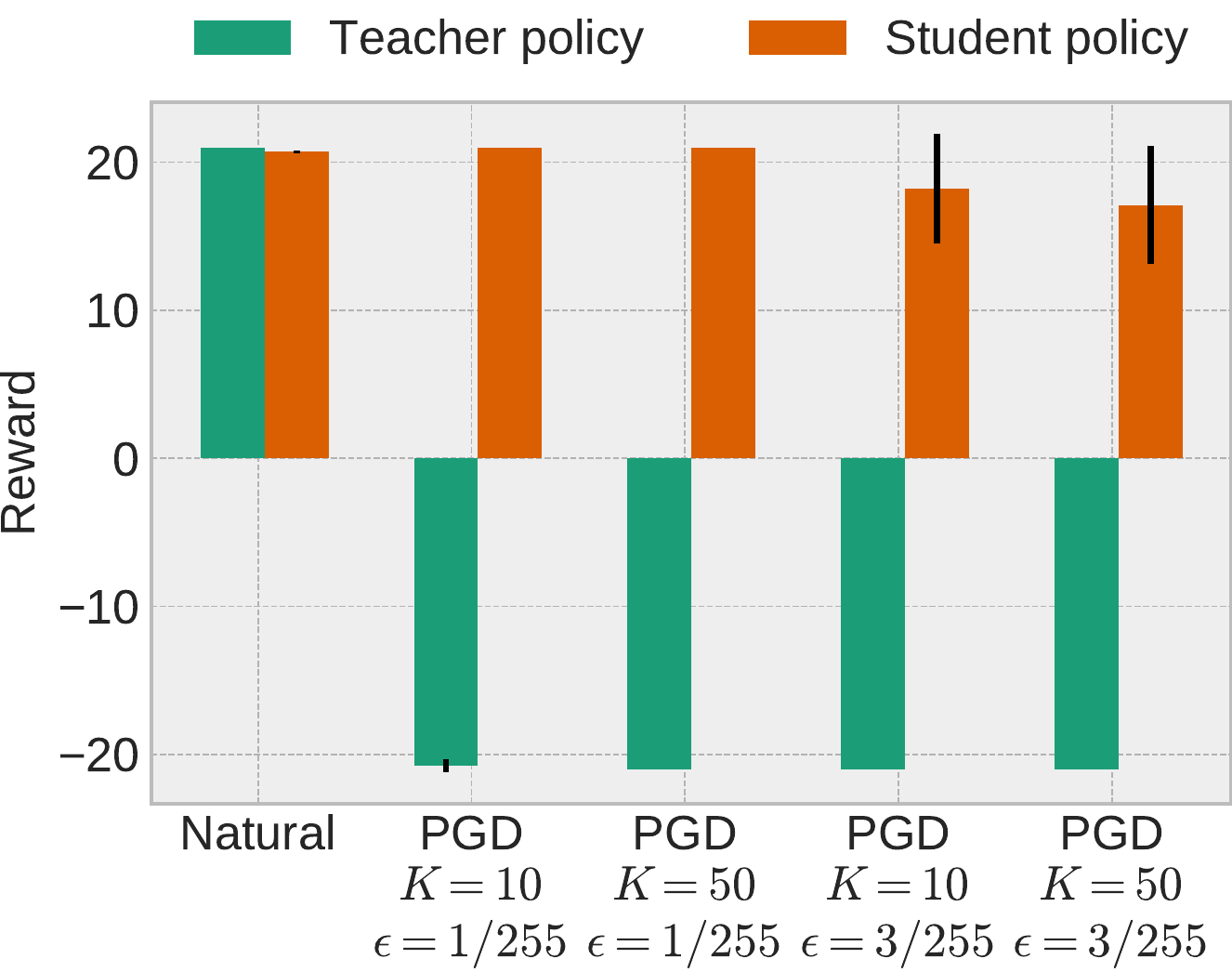}
        \caption{Pong}
    \end{subfigure}%
    \begin{subfigure}{0.33\linewidth}
        \centering
        \includegraphics[width=0.8\textwidth]{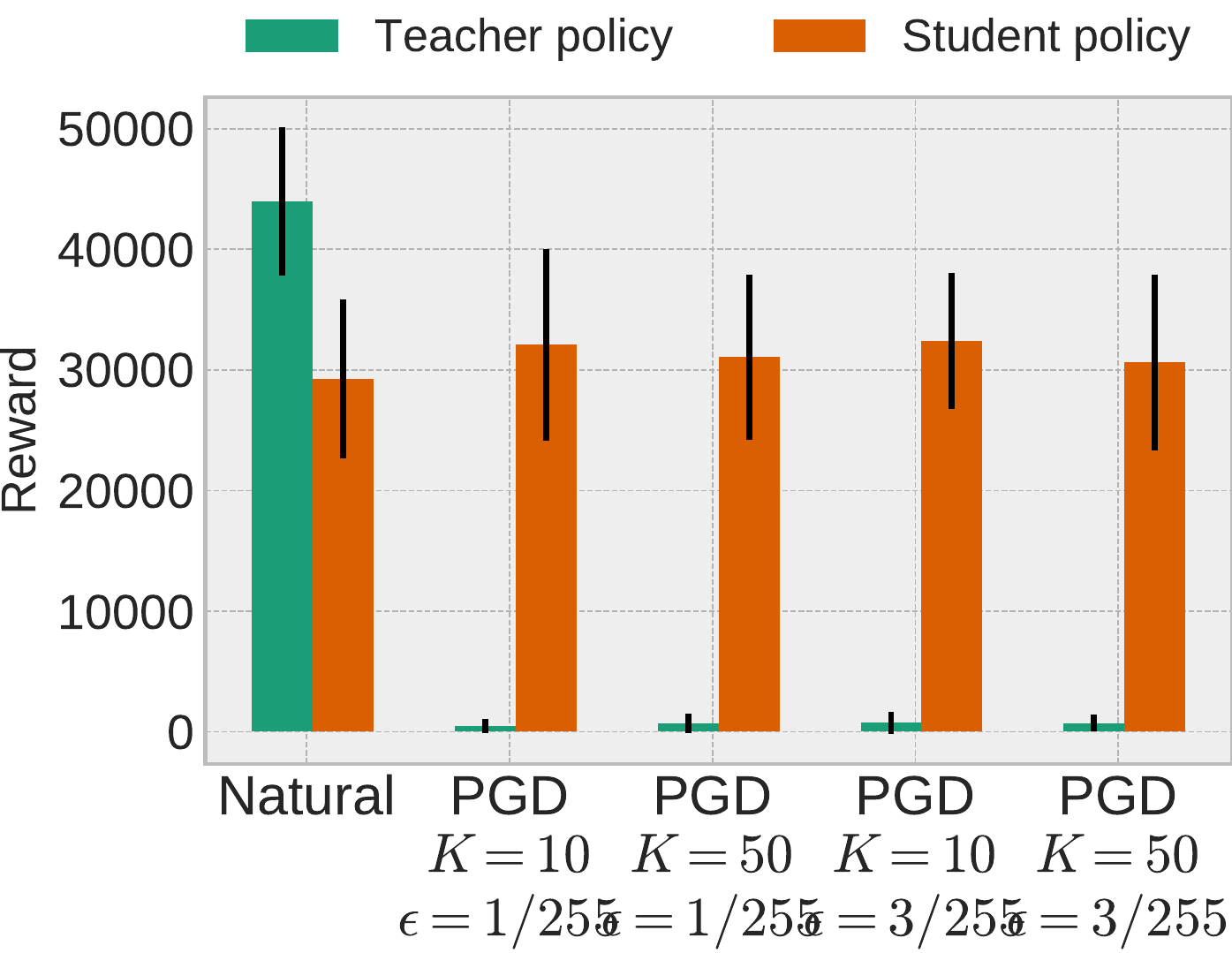}
        \caption{Road Runner}
    \end{subfigure}%
    \caption{The adversarial robustness comparison between teacher policy $\pi_{\theta^T}$ and student policy $\pi_{\theta^S}$.}
    \label{fig:teacher_student_robustness}
    \vspace{-0.2in}
\end{figure*}

In general, the accumulated rewards of our distilled policies $\pi_{\theta^S}$ under different attacks are larger than those from previous approaches; this indicates that our A2PD achieves significant robustness improvements although without adversary involved in the training.
Specifically, several interesting findings are noted. 
\textbf{(1)} Under PGD $(K=10,\epsilon=1/255)$ attack (i.e., shaded by \colorbox{blue!10}{light blue}), our $\pi_{\theta^S}$ achieves a much higher reward than all the start-of-the-arts. Especially on Bank Heist, our A2PD achieves the average reward of 1632.4, whilst in previous studies the corresponding reward values are less than 1250. Another noteworthy finding is on Freeway, where our A2PD achieve the reward $34.0\pm0.0$, the maximum that any policy can get even without attack.
On Road Runner, the reward of our A2PD is smaller than that of SA-DQN (convex), which however is mainly caused by the fact that our natural reward $29252.0\pm6610.7$ is significantly smaller than $44638.0\pm7367.0$ in SA-DQN. Further comparison and illustration of this issue (i.e., based on the relative robustness) is provided in Table~\ref{tab:relative_robustness}.
\textbf{(2)}
The similar superiority of our A2PD can also be found in PGD $(K=50,\epsilon=1/255)$ attack (i.e., shaded by \colorbox{brown!10}{light brown}) versus SA-DQN (convex)~\cite{zhang2020robust}, where our A2PD outperforms SA-DQN (convex) on 4/5 games. On Road Runner, the natural reward of SA-DQN is much larger; thus making sense that our A2PD performs slightly worse under attack. 
\textbf{(3)} In addition to $K\in\{10,50\}$ with $\epsilon=1/255$, we further evaluated our A2PD under 3 times stronger PGD attacks with $\epsilon=3/255$. According to Table \ref{table:robustness comparison}, the policies learned by our A2PD still behave robustly. For instance, the policy learned on Bank Heist achieves the reward of $1622.0\pm35.0$, which is still slightly larger than the corresponding natural reward $1617.4\pm34.8$. \textbf{(4)} Besides the PGD attack, we also evaluate the policies trained by our A2PD on FGSM and JSMA, which further showcases the ability of A2PD for achieving adversarial robustness. \textbf{(5) Relative robustness $\mathcal{M}(\mathcal{R}_\delta,\mathcal{R}_{w/o\ \delta}):$} In most of previous studies, only the reward under PGD $(K=10,\epsilon=1/255)$ attack is evaluated, but the corresponding baselines (i.e., natural reward) are different. Therefore, we further compare the relative robustness defined in Eq. (\ref{eq:relative_robustness}). The comparison shown in Table \ref{tab:relative_robustness} indicates that on all the games, our A2PD achieves the highest relative robustness. More importantly, the rewards from our A2PD are greater than or equal to 100\% on all the five games.
In summary, all the aforementioned robustness of our approach is achieved without leveraging any knowledge of adversaries; this thereby provides a totally new routine for achieving robustness in RL.

\subsection{Robustness Comparison between Teacher Policy $\pi_{\theta^T}$ and Student Policy $\pi_{\theta^S}$}
In this part, we compare the robustness between our own teacher policy $\pi_{\theta^T}$ and student policy $\pi_{\theta^S}$. To this end, we evaluate both  $\pi_{\theta^T}$ and  $\pi_{\theta^S}$ under four PGD attacks (i.e., $K\in\{10, 50\}$ and $\epsilon\in\{1/255,3/255\}$).
The results are depicted in Fig. \ref{fig:teacher_student_robustness}. 
As a whole, the rewards of teacher policies decrease significantly under attacks, whereas the rewards of student policies keep remarkably robust under attacks. 
For instance, in Fig.~\ref{fig:teacher_student_robustness} (c) on Freeway and Fig.~\ref{fig:teacher_student_robustness} (e) on Road Runner, all the rewards of teacher policy under attacks are decreased to nearly 0 (i.e., the bar is hardly observable). Similarly, in Fig.~\ref{fig:teacher_student_robustness} (d) on Pong, the rewards under attacks decrease to -21 that is the minimum reward value any policy could achieve.
\textit{In contrast}, on Freeway, Bank Heist and Pong, we observe that the student policies achieve approximately same rewards with their teacher policies and keep stable under attacks; this suggests that our distillation can achieve robustness without sacrificing performance on rewards. Although on Boxing and Road Runner the student policies cannot get rewards as high as their teacher policies, they still perform significantly more robust than the corresponding teacher policies. This firmly shows that our A2PD dramatically enhances the robustness even the data is collected from non-robust teacher policies.

\begin{figure*}[t]
    \centering
    \begin{subfigure}{0.45\linewidth}
        \centering
        \includegraphics[width=\textwidth]{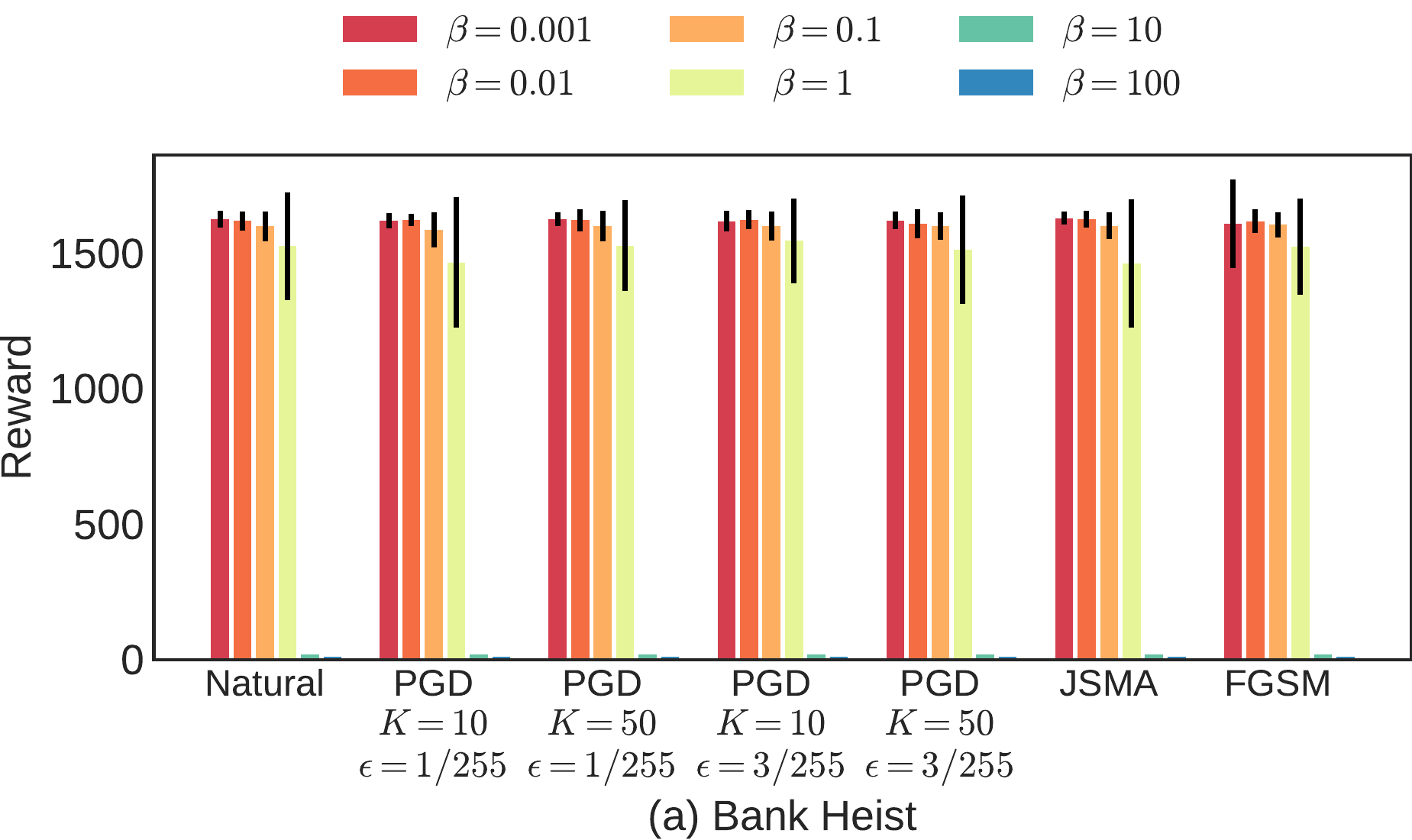}
    \end{subfigure}%
    \hspace{0.1in}
    \begin{subfigure}{0.45\linewidth}
        \centering
        \includegraphics[width=\textwidth]{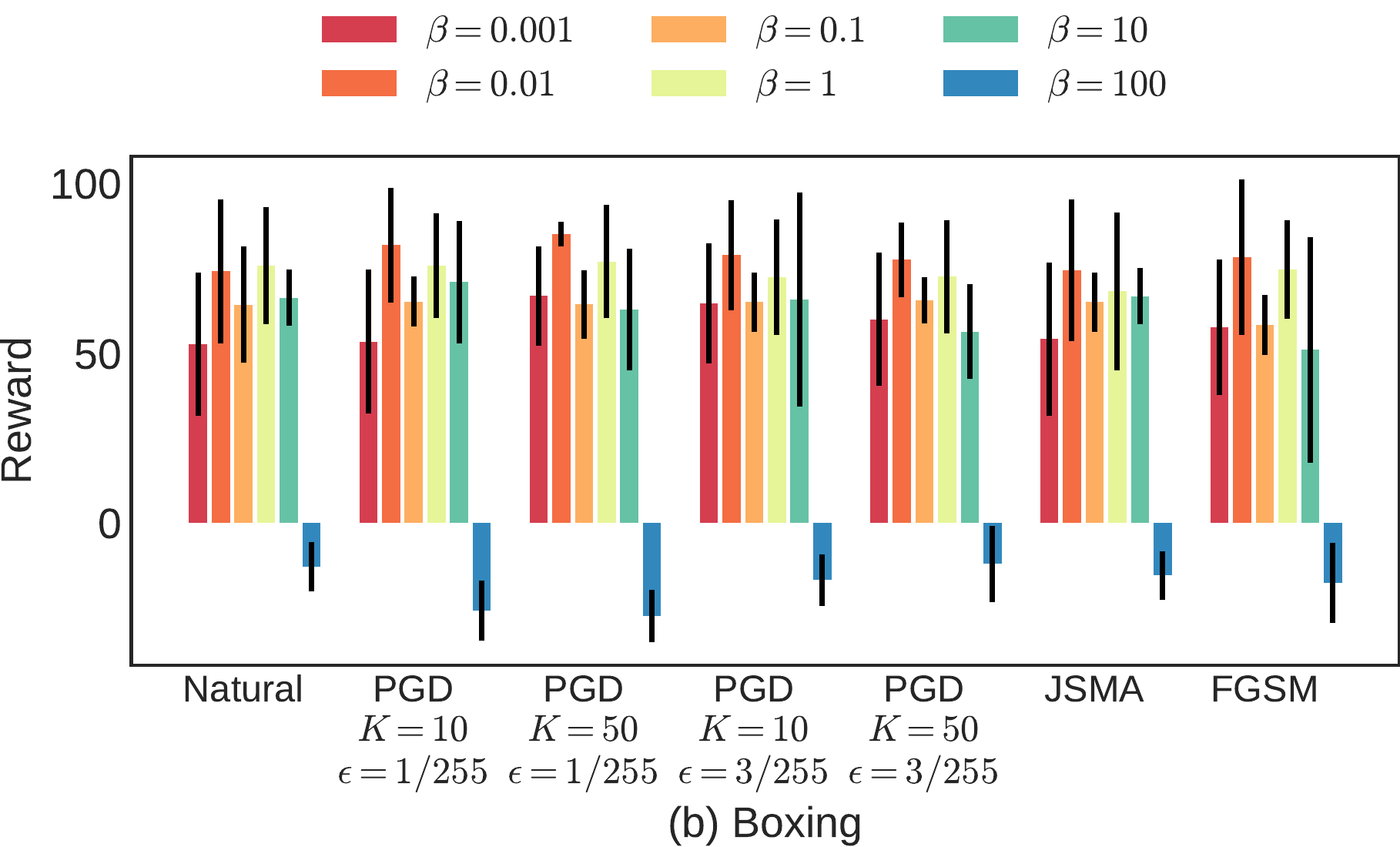}
    \end{subfigure}%
    \\
    \begin{subfigure}{0.45\linewidth}
        \centering
        \includegraphics[width=\textwidth]{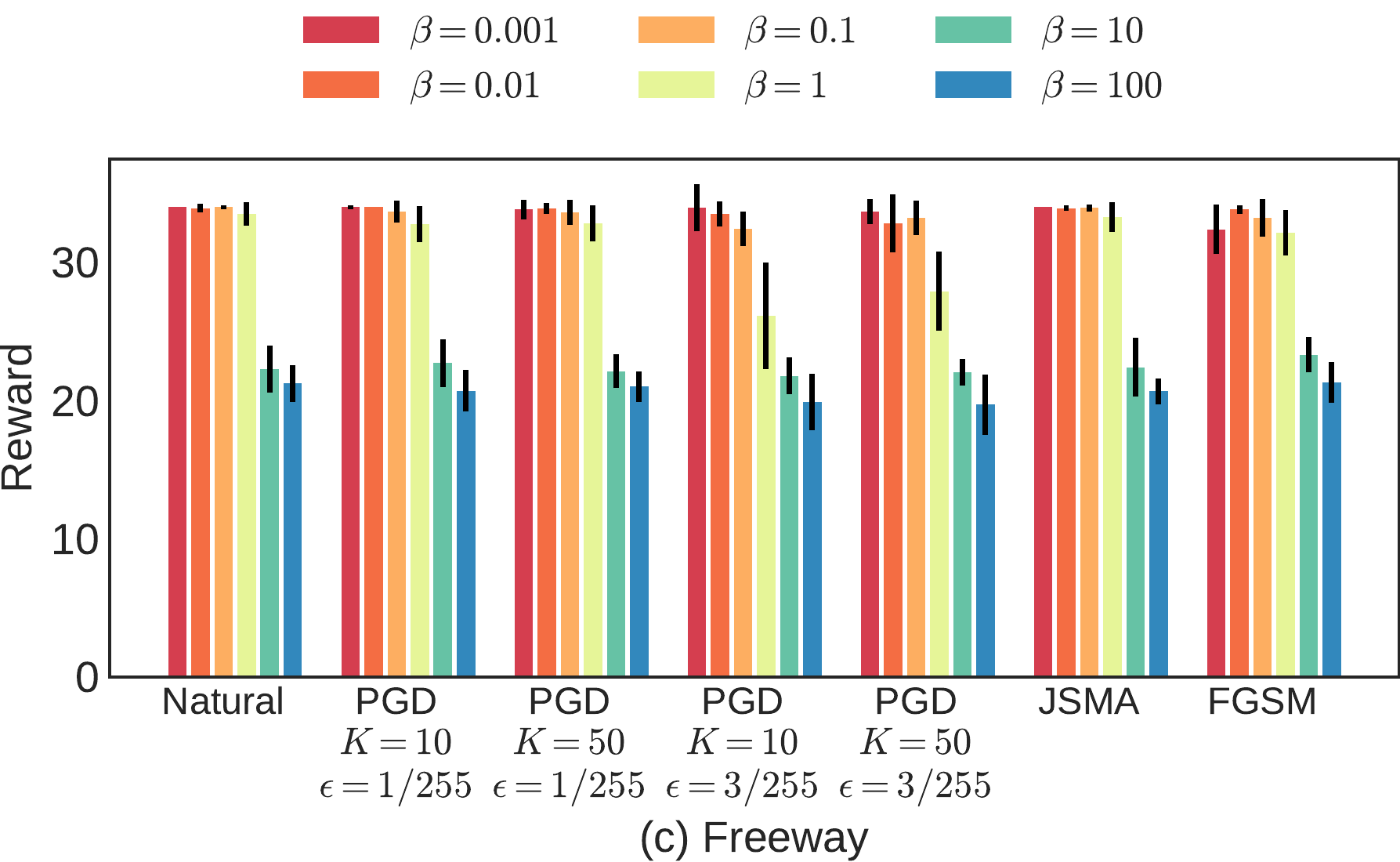}
    \end{subfigure}%
    \hspace{0.1in}
    \begin{subfigure}{0.45\linewidth}
        \centering
        \includegraphics[width=\textwidth]{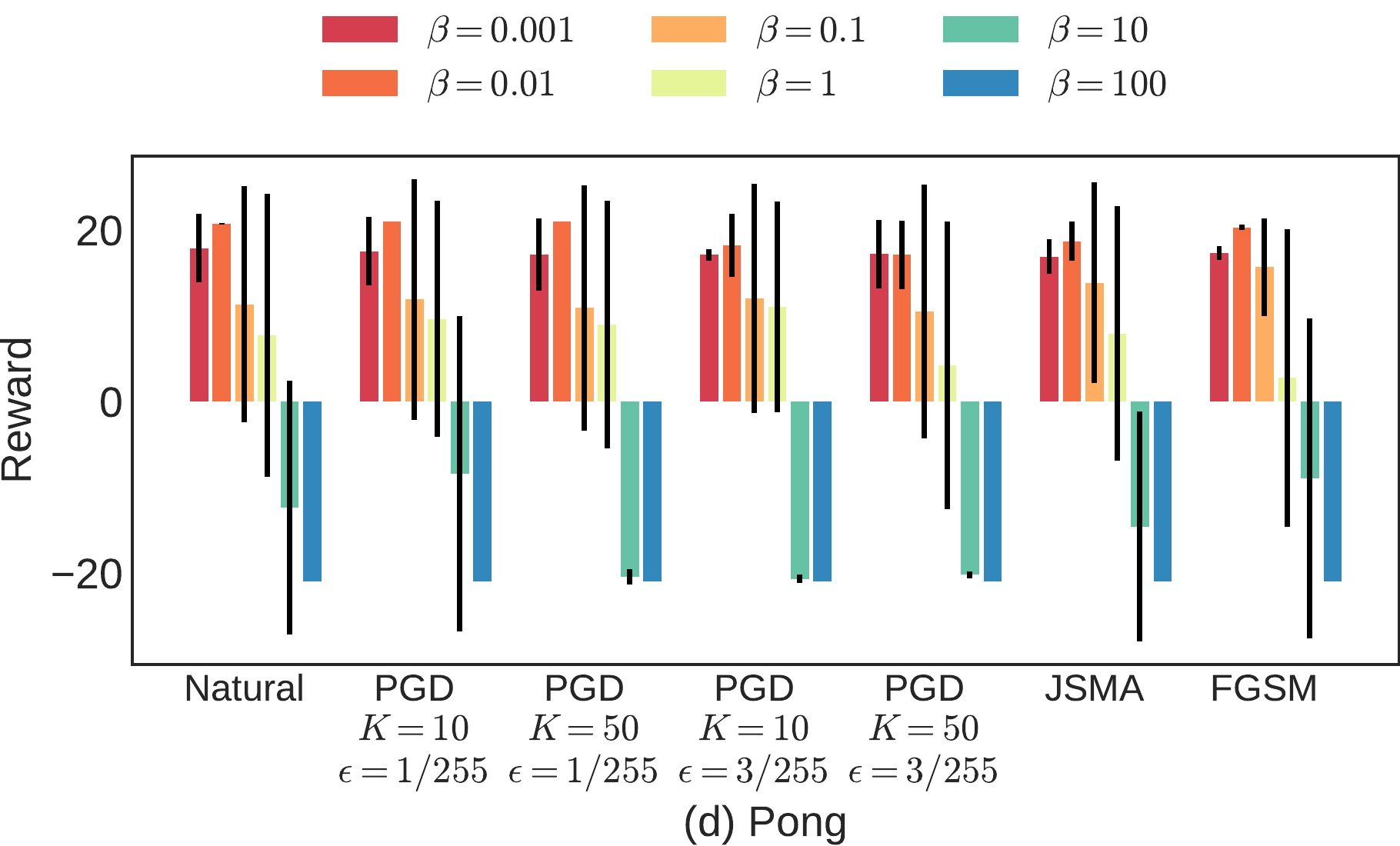}
    \end{subfigure}%
    \\
    \begin{subfigure}{0.45\linewidth}
        \centering
        \includegraphics[width=\textwidth]{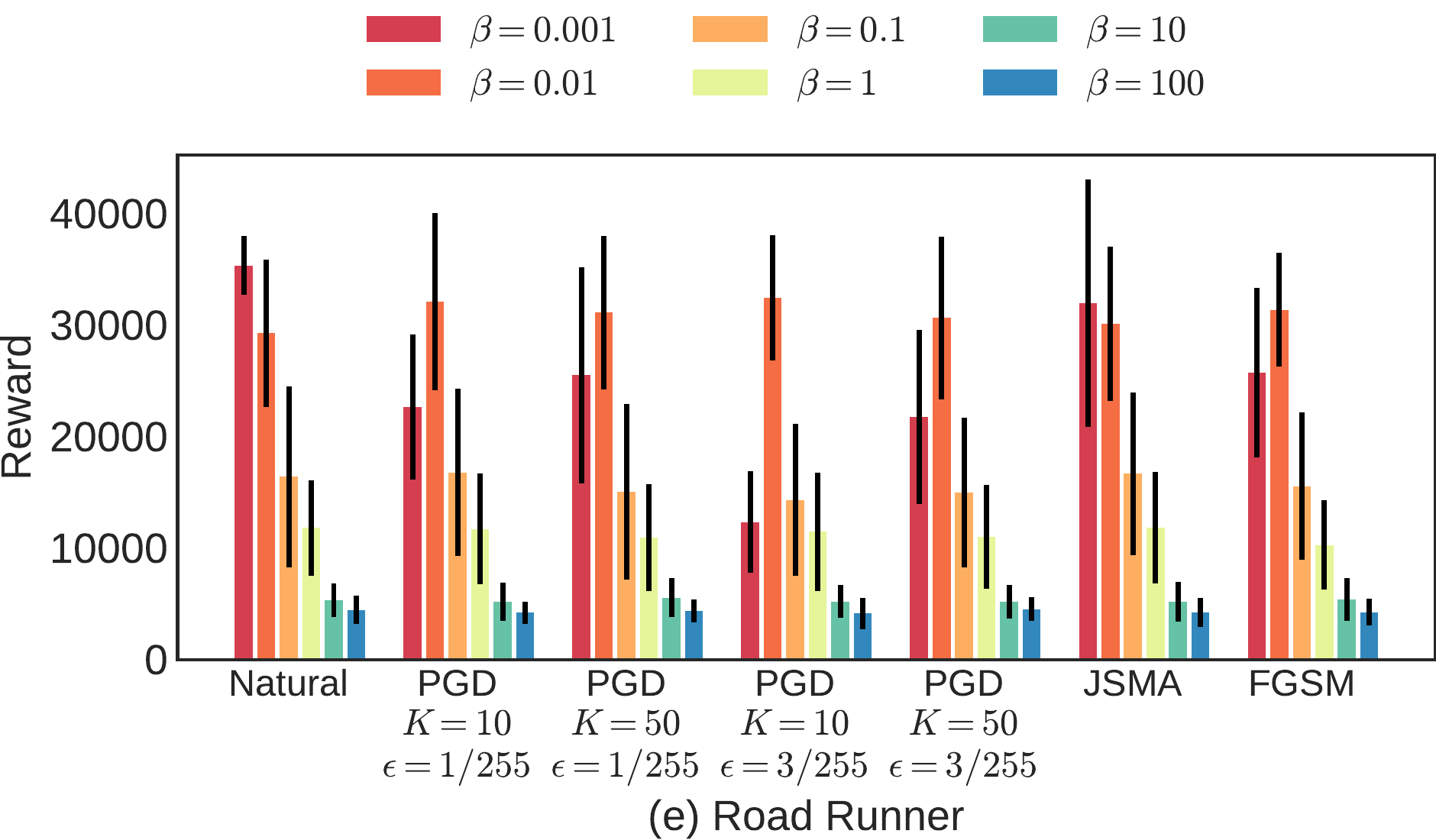}
    \end{subfigure}%
    \caption{The analysis on hyperparameter $\beta$. With each $\beta$, the distilled policy is evaluated for 50 episodes to calculate the mean (i.e., the height of the histogram) and standard deviation (i.e., the error bar on each histogram).}
    \label{fig:beta_analysis}
\end{figure*}
\begin{figure*}[t]
    \centering
    \includegraphics[width=\textwidth]{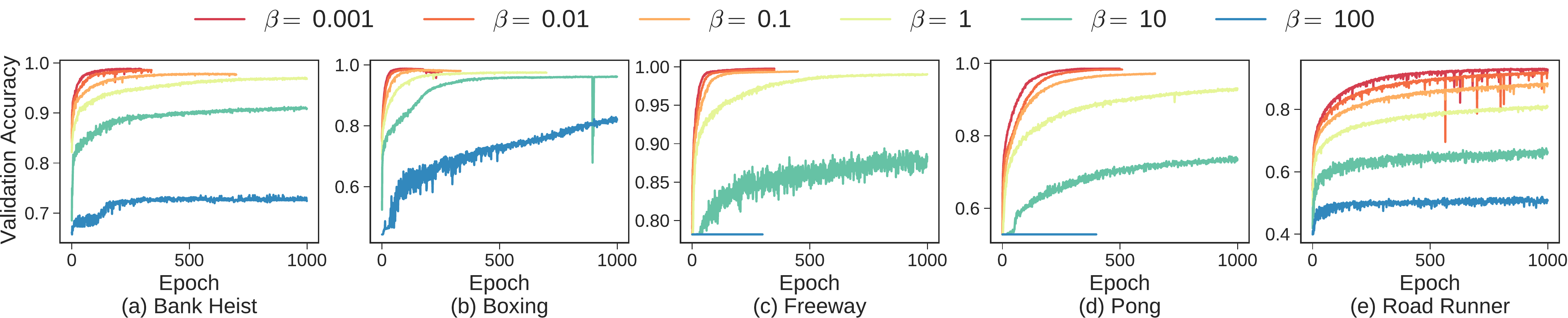}
    \caption{The convergence of A2PD training with different $\beta$. }
    \label{fig:beta_analysis_convergence}
    \vspace{-0.2in}
\end{figure*}

\subsection{Analysis on Hyper-parameter $\beta$}
We analyze the impact of $\beta$ in Eq.~(\ref{eq:pd loss}), which controls the weight of JR loss. In particular, we test six different values for $\beta$  (i.e., $\beta\in\{0.001, 0.01, 0.1, 1, 10, 100\}$) on the five Atari games. The results are shown in Fig.~\ref{fig:beta_analysis}. In general, we observe that on most cases the best performance is achieved with $\beta=0.01$. Moreover, when $\beta$ is too large, the rewards (with and without attacks) of distilled policies decrease accordingly.
These empirical observations are consistent with our analysis in Theorem~\ref{theorem:3}, where the adversarial robustness is based on maximizing the prescription gap even with adversaries in state observations. Therefore, a too large $\beta$ will overemphasize the contribution of the JR loss, thus impacting the optimization towards PGM loss that controls the accuracy of imitating the teacher policy's action selection $a^T$. 

In particular, on Bank Heist, the rewards of the distilled policies keep stable when $\beta\leq1$, while dramatically decreases to nearly 0 with $\beta\in\{10, 100\}$; thus the corresponding bars are invisible in Fig. \ref{fig:beta_analysis}(a). In Fig. \ref{fig:beta_analysis}(b), the highest reward under attack is achieved when $\beta=0.01$ and decreases to negative values when $\beta=100$. On Freeway, compared to other games, the robustness achieved is relatively less sensitive to the value of $\beta$, with a slight reward decrease when $\beta\geq 10$.
The opposite situation occurs on Pong and Road Runner. 
On Pong, when $\beta=100$, the natural reward and the rewards under attacks are -21. On Road Runner, the distilled policy with $\beta=0.01$ keeps robust for all the attacks, while other values of $\beta$ cannot achieve such consistent robustness; such comparison is most obvious on PGD ($K=10, \epsilon=3/255$) attack.

Moreover, we also evaluate the impact of $\beta$ on the convergence process of A2PD training; this further unveils the relationship between our achieved performance and the varying value of $\beta$. The results are shown in Fig. \ref{fig:beta_analysis_convergence}, from which we find that in general the A2PD training convergence becomes slower with the increasing of $\beta$. This finding is also consistent with our theoretical analysis, viz., the $\beta$ only controls the Jacobian minimization. Therefore, when $\beta$ is too large, it hurts the optimization towards PGM loss (i.e., responsible for the accuracy of A2PD training). In particular, on most games (i.e., Bank Heist, Freeway, Pong and Road Runner), when $\beta\geq10$, there are significant validation accuracy drops. Especially with $\beta=100$, the validation accuracy on Road Runner is even less than 50\%. Contrarily, the validation accuracy on most games gradually converges to the optimum if $\beta\leq0.1$.

\begin{figure*}[t]
    \centering
    \includegraphics[width=\textwidth]{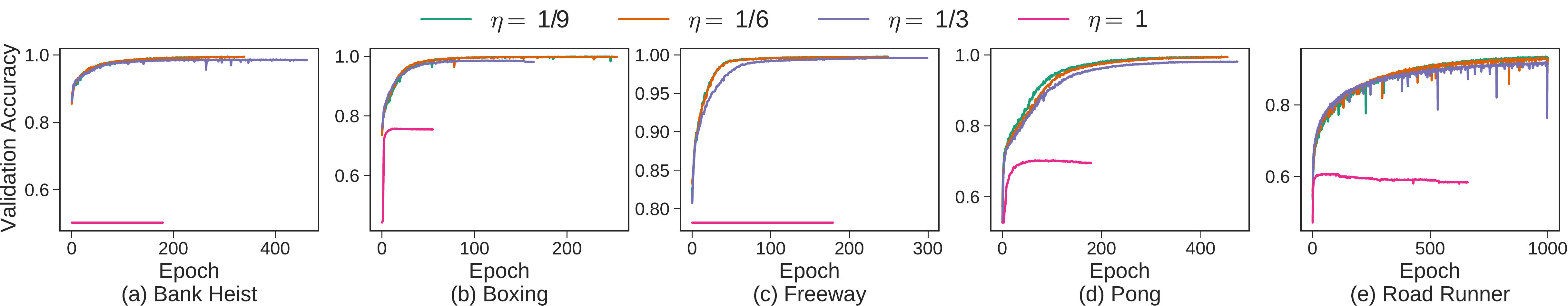}
    \caption{The convergence of A2PD training with different $\eta$ in Eq. (\ref{eq:pgm}).}
    \label{fig:alpha_analysis_convergence}
    \vspace{-0.2in}
\end{figure*}
\begin{figure*}[t]
    \centering
    \includegraphics[width=1\textwidth]{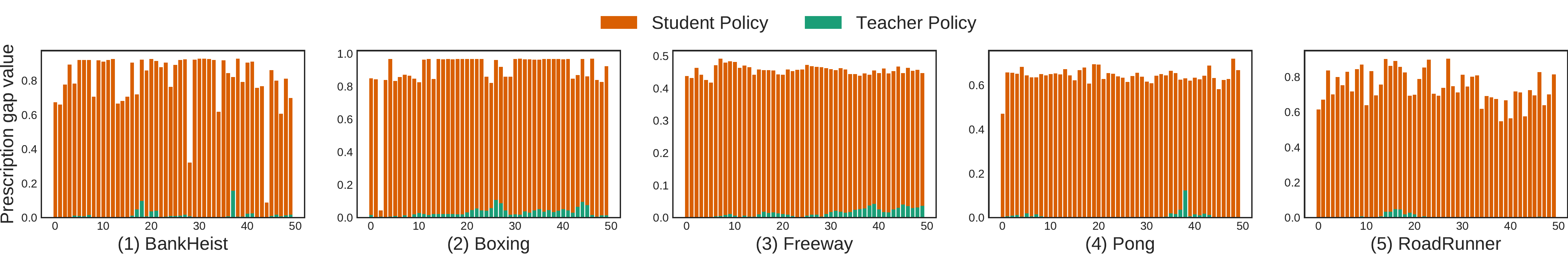}
    \caption{Comparison of prescription gap between teacher policy $\pi_{\theta^T}$ and student policy $\pi_{\theta^S}$
    }\label{fig:prescription_compare}
\end{figure*}

\subsection{Analysis on Hyperparameter $\eta$ in $\mathcal{L}_{pgm}(\theta^S)$}
In Eq. (\ref{eq:pgm}), the hyperparameter $\eta$ is designed to balance the maximization of $\pi_{\theta^S}(s_t,a^T)$ and the entropy regularization; thus it is expected to impact the convergence of A2PD training. Therefore, we further analyze the hyperparameter $\eta$ with evaluating different values from $\{1/9, 1/6, 1/3, 1\}$. In the training, we utilize early stop with setting patience as 60. The maximum epoch value is 1000. The convergence plot on the five games are shown in Fig. \ref{fig:alpha_analysis_convergence}. From the results, we can generally see that the convergence becomes faster with the decreasing of $\eta$. In particular, the convergence processes with $\eta\leq1/3$ keep similar, while the convergence with $\eta=1$ got stuck. This suggests that the maximization of $\pi_{\theta^S}(s_t,a^T)$ and the entropy regularization in Eq. (\ref{eq:pgm}) can be well balanced with the setting $\eta\in[1/9,1/3]$ on Atari games. 

\subsection{Analysis on Prescription Gap}
To examine our claim on improving adversarial robustness via maximizing the prescription gap between optimal action and other sub-optimal actions, we compare the prescription gap between the teacher policy $\pi_\theta^T$ and student policy $\pi_\theta^S$ as shown in Fig.~\ref{fig:prescription_compare}. Assuming a prescription distribution $P=[p_0, p_1,\cdots, p_N]$, the prescription gap is calculated as $p_{i^*} -\max_{j\neq i^*}p_j, \text{ where }i^*=\argmax_i p_i$. From Fig.~\ref{fig:prescription_compare}, on all the five games tested, the prescription gap values of student policies are far larger than those of teacher policies. Together with the above robustness analysis, it further demonstrates that the robustness of DRL policies is indeed improved via maximizing the prescription gap based on our policy distillation loss in Eq.~(\ref{eq:pd loss}). Most importantly, in achieving the adversarial robustness for DRL policies by maximizing the prescription gap, no knowledge of the adversaries is required, which thus enables a more realistic solution for improving adversarial robustness compared to adversarial training based approaches.

\subsection{Analysis on Computational Efficiency} 
To support our claim about computational efficiency, we compare the averaged training time of our A2PD and PGD based adversarial training. Note that, the maximum epoch and patience for early stop are set same for all the student policy training, thus we can compare the averaged training time on each epoch. The statistics of training time is obtained on NVIDIA RTX 2080 Ti. The results are shown in Table~\ref{tab:computation_time}, where the average training time (seconds) from 50 epochs on the five games are presented. In general, we can see that our A2PD requires significantly less computation time than all the adversarial training approaches. Specifically, in average on all games, our A2PD only requires 27\% of the training time in PGD-20 adversarial training. 

\begin{table}[t]
    \centering
    \scriptsize
    \renewcommand{\arraystretch}{1.6}
    \caption{Computation time comparison on five games. The value is the average over 50 epochs. The unit of time is second.}
    \begin{tabular}{l|cccc}
    \specialrule{.15em}{.05em}{.05em}
    \multirow{2}{*}{Time on Games} &\multirow{2}{*}{A2PD} &\multicolumn{3}{|c}{Adversarial Training} \\
    \cline{3-5}
    & & \multicolumn{1}{|c}{PGD-4}&PGD-10 & PGD-20 \\
    \specialrule{.1em}{.05em}{.05em}
    \textit{Time (Bank Heist)} & $\textbf{18.20} s$ & \multicolumn{1}{c}{$23.29 s$} & $36.89 s$ & $64.76 s$  \\
    \textit{Time (Boxing)} & $\textbf{17.23} s$ & \multicolumn{1}{c}{$25.73 s$} & $40.12 s$ & $67.32 s$  \\
    \textit{Time (Freeway)} & $\textbf{19.73} s$ & \multicolumn{1}{c}{$26.72 s$} & $43.54s$ & $68.57 s$  \\
    \textit{Time (Pong)} & $\textbf{19.34} s$ & \multicolumn{1}{c}{$26.52 s$} & $41.47 s$ & $68.36 s$  \\
    \textit{Time (Road Runner)} & $\textbf{17.41} s$ & \multicolumn{1}{c}{$27.83 s$} & $43.26 s$ & $71.51 s$  \\
    \specialrule{.1em}{.05em}{.05em}
    \textit{Average} & $\textbf{18.38} s$ & $26.02 s$ & $41.09 s$ & $68.10 s$  \\
    \specialrule{.15em}{.05em}{.05em}
    \end{tabular}
    \label{tab:computation_time}
    \vspace{-0.2in}
\end{table}

\section{Conclusion}
For the first time, this paper proposes an adversary agnostic robust deep reinforcement learning paradigm (i.e., based on policy distillation) that achieves robust policy to resist against adversaries but without relying on any information of adversaries during training. 
To this end, we theoretically derive that the robustness of student policy can be indeed learned independent with the adversaries. Accordingly, we propose a novel PD loss function that contains: 1) a PGM loss for simultaneously maximizing the probability of the action prescribed by teacher policy as well as the entropy of unwanted actions; 2) a JR loss that minimizes the norm of Jacobian with respect to the input state. The theoretical analysis proves that our propose PD loss guarantees to increase the prescription gap and the adversarial robustness. Meanwhile, experiments on five Atari games show that the robustness of the student policies trained via our PD loss is significantly improved.

\bibliographystyle{unsrt}
\bibliography{references}
\newpage
\begin{table}[t]
    \centering
    \caption{Hyper-parameters settings for training teacher policy $\pi_{\theta^T}$.}
    \renewcommand{\arraystretch}{1.3}
    \begin{tabular}{l|l|l}
    \specialrule{.15em}{.05em}{.05em}
    \textbf{Parameters} & \textbf{Settings}  & \textbf{Descriptions}\\
    \specialrule{.15em}{.05em}{.05em}
    Optimizer & Adam &-- \\
    Batch size & 32 &-- \\
    Learning rate & 0.000625 & Adam learning rate\\
    $\phi_1$ & 0.9 & Adam decay rate 1\\
    $\phi_2$  & 0.999 & Adam decay rate 2 \\
    Adam-eps & $1.5\times 10^{-4}$ & Adam epsilon\\
    Start steps & $2\times 10^4$ & Number of steps before starting training\\
     Environment ID    & 123 & The random seed in Arcade environment\\
     T-max & $1\times 10^7$ & Number of training steps\\
     Max-episode-len & $108 \times 10^3$ & Maximum episode length in game frames 
     \\
     $h$ & 4 & Number of consecutive states processed\\
     Hidden-size & 512 & Network hidden size\\
     $\sigma$ & 0.1 & Initial standard deviation of noisy linear layers\\
     Atoms  & 51 & Discretised size of value distribution \\
     V-min & -10 & Minimum of value distribution support\\
     V-max & 10 & Maximum of value distribution support\\
     Memory-length & $1\times 10^6$ & The length of replay buffer\\
     Target-update & $1\times 10^4$ & The frequency of updating target network\\
     \specialrule{.15em}{.05em}{.05em}
    \end{tabular}
    \label{tab:parameter setting}
    \vspace{-0.15in}
\end{table}

\begin{table}[ht]
    \centering
    \caption{Hyper-parameters settings for our proposed policy distillation of $\pi_{\theta^S}$.}
    \renewcommand{\arraystretch}{1.3}
    \addtolength{\tabcolsep}{1pt}
    \begin{tabular}{l|l|l}
    \specialrule{.15em}{.05em}{.05em}
    \textbf{Parameters} & \textbf{Settings}  & \textbf{Descriptions}\\
    \specialrule{.15em}{.05em}{.05em}
    Batch size & 32 &-- \\
    Learning rate & 0.00004 & Adam learning rate\\
    $\phi_1$ & 0.9 & Adam decay rate 1\\
    $\phi_2$  & 0.999 & Adam decay rate 2 \\
    Adam-eps & $1\times 10^{-7}$ & Adam epsilon\\
    Max-epoch & 1000 & The maximum number of epochs\\
    Patience & 60 & Early stop number\\
    $\beta$ & 0.1 & The weight of $\mathcal{L}_{jr}(\theta^S)$ in Eq.~(\ref{eq:pd loss})\\
    $\eta$ & 1/3 & Discount factor in Eq.~(\ref{eq:pgm})\\
     \specialrule{.15em}{.05em}{.05em}
    \end{tabular}
    \label{tab:student_parameter_setting}
    \vspace{-0.1in}
\end{table}
\end{document}